\documentclass[11pt]{article}

\PassOptionsToPackage{table,dvipsnames}{xcolor}
\usepackage[final]{acl}

\makeatletter
\AtBeginDocument{%
  \RemoveFromHook{build/column/before}[lineno]%
  \def\@LN@makecol{%
    \@LN@orig@makecol
    \setbox\@outputbox \vbox{%
      \boxmaxdepth \@maxdepth
      \protected@write\@auxout{}{%
        \string\def\string\@LN@column{\if@firstcolumn1\else2\fi}%
      }%
      \box\@outputbox
    }%
  }%
}
\makeatother
\usepackage{titletoc}
\usepackage{booktabs}
\usepackage{multirow}
\usepackage{pifont}

\definecolor{LightGray}{gray}{0.96}
\definecolor{HeaderGray}{gray}{0.90}

\usepackage{times}
\usepackage{latexsym}

\usepackage[T1]{fontenc}

\usepackage[utf8]{inputenc}

\usepackage{microtype}

\usepackage{inconsolata}
\usepackage{amssymb}

\usepackage{listings}
\usepackage[most]{tcolorbox}

\usepackage{graphicx}

\definecolor{PromptNeutral}{RGB}{248,244,230}
\definecolor{ProtocolStatic}{RGB}{76,120,168}
\definecolor{ProtocolFull}{RGB}{245,133,24}
\definecolor{ProtocolPartial}{RGB}{84,162,75}
\definecolor{ProtocolOracle}{RGB}{178,121,162}

\newcommand{\staticprotocol}{\texorpdfstring{\textcolor{ProtocolStatic}{\textsc{Static}}}{Static}}
\newcommand{\fullprotocol}{\texorpdfstring{\textcolor{ProtocolFull}{\textsc{Full}}}{Full}}
\newcommand{\partialprotocol}{\texorpdfstring{\textcolor{ProtocolPartial}{\textsc{Partial}}}{Partial}}
\newcommand{\oracleprotocol}{\texorpdfstring{\textcolor{ProtocolOracle}{\textsc{None}}}{None}}

\definecolor{PromptNeutral}{RGB}{248,248,248}

\newtcblisting{promptbox}{
    colback=PromptNeutral,
    colframe=black!35,
    boxrule=0.4pt,
    borderline={0.4pt}{0pt}{black!35,dashed},
    arc=2pt,
    left=1.2mm,right=1.2mm,top=1.2mm,bottom=1.2mm,
    breakable,
    enhanced,
    width=\textwidth,
    listing only,
    listing options={
        basicstyle=\ttfamily\footnotesize,
        breaklines=true,
        breakatwhitespace=false,
        columns=fullflexible,
        keepspaces=true,
        showstringspaces=false
    }
}

\newtcblisting{promptboxonecol}{
    colback=PromptNeutral,
    colframe=black!35,
    boxrule=0.4pt,
    borderline={0.4pt}{0pt}{black!35,dashed},
    arc=2pt,
    left=1.2mm,right=1.2mm,top=1.2mm,bottom=1.2mm,
    breakable,
    enhanced,
    width=\linewidth,
    listing only,
    listing options={
        basicstyle=\ttfamily\scriptsize,
        breaklines=true,
        breakatwhitespace=false,
        columns=fullflexible,
        keepspaces=true,
        showstringspaces=false
    }
}

\newtcblisting{staticpromptbox}{
    colback=ProtocolStatic!8,
    colframe=ProtocolStatic!70,
    boxrule=0.5pt,
    arc=3pt,
    left=1.2mm,right=1.2mm,top=1.2mm,bottom=1.2mm,
    breakable,
    enhanced,
    width=\textwidth,
    listing only,
    listing options={
        basicstyle=\ttfamily\footnotesize,
        breaklines=true,
        breakatwhitespace=false,
        columns=fullflexible,
        keepspaces=true,
        showstringspaces=false
    }
}

\newtcblisting{fullpromptbox}{
    colback=ProtocolFull!8,
    colframe=ProtocolFull!70,
    boxrule=0.5pt,
    arc=3pt,
    left=1.2mm,right=1.2mm,top=1.2mm,bottom=1.2mm,
    breakable,
    enhanced,
    width=\textwidth,
    listing only,
    listing options={
        basicstyle=\ttfamily\footnotesize,
        breaklines=true,
        breakatwhitespace=false,
        columns=fullflexible,
        keepspaces=true,
        showstringspaces=false
    }
}

\newtcblisting{partialpromptbox}{
    colback=ProtocolPartial!8,
    colframe=ProtocolPartial!70,
    boxrule=0.5pt,
    arc=3pt,
    left=1.2mm,right=1.2mm,top=1.2mm,bottom=1.2mm,
    breakable,
    enhanced,
    width=\textwidth,
    listing only,
    listing options={
        basicstyle=\ttfamily\footnotesize,
        breaklines=true,
        breakatwhitespace=false,
        columns=fullflexible,
        keepspaces=true,
        showstringspaces=false
    }
}

\newtcblisting{oraclepromptbox}{
    colback=ProtocolOracle!8,
    colframe=ProtocolOracle!70,
    boxrule=0.5pt,
    arc=3pt,
    left=1.2mm,right=1.2mm,top=1.2mm,bottom=1.2mm,
    breakable,
    enhanced,
    width=\textwidth,
    listing only,
    listing options={
        basicstyle=\ttfamily\footnotesize,
        breaklines=true,
        breakatwhitespace=false,
        columns=fullflexible,
        keepspaces=true,
        showstringspaces=false
    }
}

\newcommand{\owen}[1]{}

\title{When Seeing Is Not Enough:\\Benchmarking Interactive Visual Grounding in LVLMs}

\author{Zhengxiang Wang \\            Department of Linguistics \& IACS \\
  Stony Brook University \\
  \texttt{zhengxiang.wang@stonybrook.edu} \\\And
  Owen Rambow \\
  Department of Linguistics \& IACS \\
  Stony Brook University \\
  \texttt{owen.rambow@stonybrook.edu} \\}

\begin{document}
 \maketitle
\begin{abstract}

Visual grounding is typically evaluated as a one-shot mapping from an informative referring expression to a visual target. This formulation misses a central property of real-world reference: initial referring expressions are often incomplete or ambiguous, requiring participants to establish shared understanding through interaction. We introduce a controlled evaluation framework for interactive visual grounding in large vision-language models (LVLMs), varying how much target information is provided upfront and how much must be acquired through dialogue. Across four human-grounded visual contexts and four interaction protocols, current LVLMs perform significantly below task-level human baselines. Interaction can help when follow-up questions refine or repair an initial target description. Performance is lowest when no initial description is provided and target information must be acquired through questions, indicating that proactive question-driven grounding remains difficult. LVLMs are also poorly calibrated, often reporting confidence that exceeds their empirical accuracy. Follow-up studies confirm these patterns across varied description sources (human versus AI), reasoning efforts, repeated interactions, description providers, and visual contexts. Overall, interactive visual grounding remains challenging, requiring visual matching, information seeking and synthesis.\footnote{GitHub project: \url{https://github.com/jaaack-wang/benchmarking-interative-visual-grounding}.}

\end{abstract}

\section{Introduction}\label{sec:introduction}

Visual grounding is a core vision-language capability that requires a system to identify and localize regions or objects in a visual environment from a textual description \cite{Xiao2026}. It underlies applications such as visual question answering, visual retrieval, human--AI interaction, and embodied AI agents. Enabled by large-scale cross-modal training and alignment \cite{pmlr-v139-radford21a,alayrac2022flamingo,llava2023}, recent large vision-language models (LVLMs) have demonstrated strong general-purpose vision-language capabilities \cite{CogVLM2024,openai2024gpt4technicalreport,geminiteam2025geminifamilyhighlycapable}; in particular, they have largely saturated standard visual grounding benchmarks \cite{chen2024revisitingreferringexpressioncomprehension,dong2026refadv}.

\begin{figure}[]
    \centering    \includegraphics[width=1\linewidth]{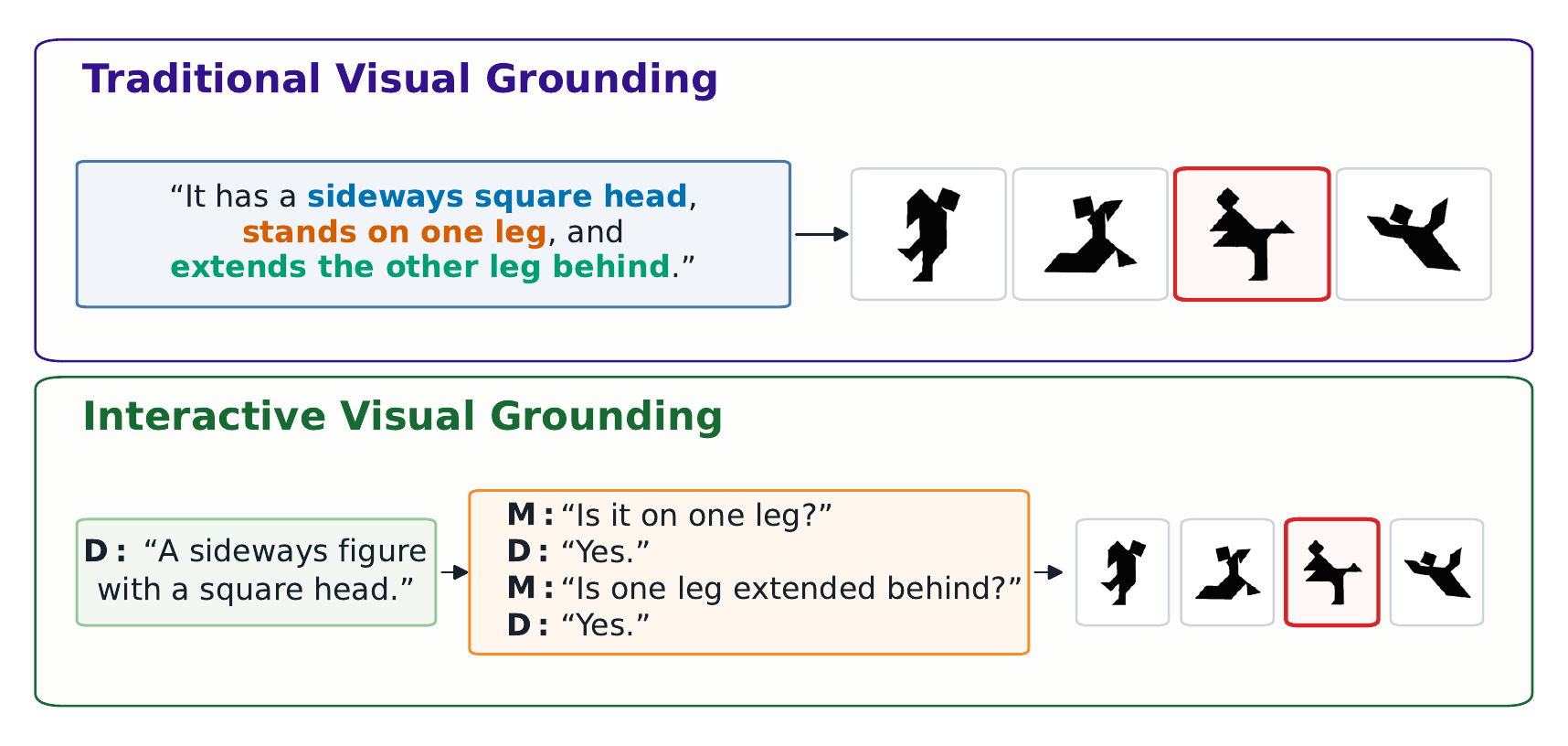}
    \caption{Traditional visual grounding performs one-shot matching, assuming an informative input description. Interactive visual grounding removes this assumption, requiring the Matcher (M) to interact with the Director (D) to seek more information. Examples are adapted from \citet{Hawkins2020} to be illustrative.}
    \label{fig:spirit-figure}
\end{figure}

However, traditional visual grounding (Figure~\ref{fig:spirit-figure} upper part) relies on a restrictive assumption: the model is given a sufficiently informative referring expression before grounding begins. Most benchmarks therefore cast grounding as a one-shot, non-interactive task, where the model maps an already available description to a visual target \cite{Mao2016}. This setup primarily evaluates \textit{static visual matching}. It does not test whether a model can recognize that a description is incomplete, ask for missing information, or update its grounding decision as new evidence becomes available.


This assumption departs from how reference works in real-world human communication (Figure~\ref{fig:spirit-figure}, lower part). Referring expressions are frequently underspecified, ambiguous, or sufficient only relative to a particular visual context. Speakers and listeners establish reference collaboratively, using clarification questions and feedback to build common ground over time \cite{Clark1986,Brennan1996}. 


We therefore study \textit{interactive visual grounding}, where target information may be ambiguous or underspecified at first and must be established through interaction. We argue that interactive visual grounding requires two broad capabilities. First, models need \textit{visual matching} ability: mapping a referring expression, or an accumulated target representation, to the intended visual item. Second, models need \textit{information seeking and synthesis} ability: assessing whether the current information is sufficient, asking effective questions when it is not, and synthesizing answers across turns. Traditional grounding benchmarks primarily test the former; interactive grounding requires both. 


Prior studies evaluate non-interactive referring-expression comprehension or object matching \cite{tang-etal-2024-grounding,wang-etal-2025-lvlms}, and interactive reference tasks that involve human--AI communication \cite{hua2024talk,imai-etal-2025-measuring,zeng2026lvlmshumansgrounddifferently,jones2026llmspeoplelearnform}. These studies reveal important limitations, but without systematically controlling whether target information is supplied or must be acquired, they cannot disentangle visual-matching limitations from failures of interactive information seeking. Consequently, aggregate performance can mask fundamentally different failure modes and offer little guidance for model improvement.

To address this issue, we introduce a controlled evaluation framework for interactive visual grounding. Our framework varies how much target information is available upfront and how much must be acquired through dialogue, thereby testing whether models can ground informative descriptions, use optional clarification, recover from underspecification, and construct target representations through their own questions. Across 38,848 target selections from eight LVLMs under diverse experimental conditions, current models perform significantly below task-level human baselines even when additional target information is available. They can use follow-up questions to refine or repair an initial description, but struggle to ground targets through proactive question-driven dialogue despite asking the most questions. Their reported confidence also frequently exceeds empirical accuracy. Together, these findings show that interactive visual grounding remains an important challenge for LVLMs beyond static visual matching.

\section{Related Work\label{sec:relatedWork}}

\paragraph{Static Visual Grounding}
Most visual grounding benchmarks evaluate grounding as a static identification or localization problem: given an image and a referring expression, a model selects the corresponding region or object~\cite{Lin2014,kazemzadeh-etal-2014-referitgame,Flickr30k2015,yu2016modelingcontextreferringexpressions,chen2024revisitingreferringexpressioncomprehension,dong2026refadv}. This formulation has been central to measuring visual-language alignment, but it assumes that the referring expression is already sufficiently informative. As a result, it primarily tests visual matching, rather than whether a model can recognize missing information or seek clarification.

\paragraph{Interactive Visual Grounding}
Prior work has introduced interaction into visual grounding through goal-oriented visual dialogue and clarification-based object identification. GuessWhat?! frames object identification as a single-target game in which a questioner asks yes/no questions to identify an object~\cite{Vries2017CVPR}. INGRESS studies situated referring-expression grounding, where a robot asks clarification questions when a referent is ambiguous~\cite{INGRESS2020}. InViG benchmarks open-ended multi-turn grounding for object disambiguation~\cite{zhang2023invigbenchmarkinginteractivevisual}. These settings show that interaction is important for grounding, but they typically focus on single-target identification or constrained question-answering. In contrast, we study multi-target grounding under controlled information conditions, allowing us to distinguish visual matching from information-seeking ability. 

\begin{figure*}[htb]
    \centering    \includegraphics[width=1\linewidth]{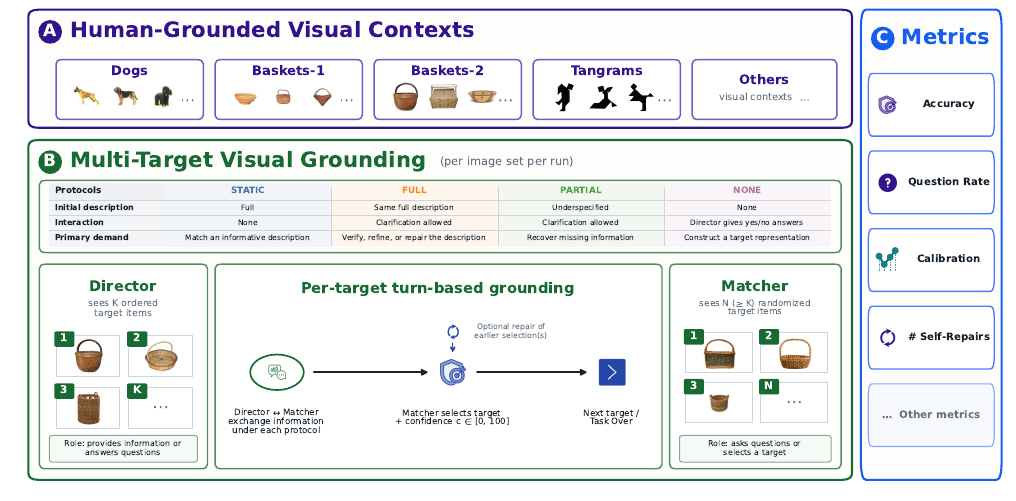}
    \caption{Overview of our proposed framework for controlled interactive visual grounding evaluation. (A) and (C) show the visual contexts and metrics used in our main experiments, respectively; (B) summarizes the interaction protocols and grounding process. The ``Others'' entries in (A) and (C) highlight the framework's extensibility.}
    \label{fig:methodology}
\end{figure*}

\paragraph{LVLMs in Reference Tasks}
Recent studies have evaluated LVLMs in repeated, multi-target reference tasks, building on classic work on referential communication and common ground~\cite{Krauss1969,Clark1986,Schober1989}. Some work examines whether LVLMs can comprehend referring expressions in human-human dialogues that become more efficient over repeated interaction~\cite{wang-etal-2025-lvlms}. Other work studies LVLM behavior in self-play~\cite{hua2024talk,imai-etal-2025-measuring} or human--AI interaction~\cite{zeng2026lvlmshumansgrounddifferently,jones2026llmspeoplelearnform}, revealing non-human-like patterns in both generation and comprehension. Our work differs by systematically controlling how much target information is provided upfront and how much must be acquired through dialogue. This lets us test whether model failures arise from visual matching or information seeking or their combination.

\paragraph{Multi-Turn Evaluation} 


Multi-turn evaluation has been increasingly used to evaluate LLMs beyond isolated single-turn prompts \cite{zheng2023judging,bai-etal-2024-mt, kwan-etal-2024-mt, wang-etal-2025-llms-perform,han-etal-2025-language}. Recent work shows that LLMs can degrade across turns and often fail to seek clarification, causing misunderstandings to compound over interaction~\cite{laban2026llms,shaikh-etal-2025-navigating}. Multimodal multi-turn evaluation extends these challenges to LVLMs, where models must track visual inputs, dialogue history, and evolving user intent~\cite{liu2024mmdu,lee2025multiverse,xue-etal-2025-mmrc,feng-etal-2023-mmdialog,tian2025chatterbox}. Our work instead tests whether models can use dialogue to determine and acquire missing target information and improve grounding accuracy.



\section{Interactive Visual Grounding}\label{sec:interactive_vg}

Figure~\ref{fig:methodology} visualizes our proposed evaluation framework, connecting the visual contexts (Panel A), multi-target visual grounding (Panel B), and evaluation metrics (Panel C). This section details the core designs for the overall task, the interaction protocols, and the evaluation metrics.

\paragraph{Task Formalization} A \emph{Director} sees an ordered target set $D=(d_1,\ldots,d_k)$; a \emph{Matcher} sees a separately indexed candidate set $M=(m_1,\ldots,m_n)$ containing the targets and possibly additional distractors. The players do not see each other's view, and indices are randomized independently.

For each target position $j$, the players exchange utterances until the Matcher selects $\hat{d}_j\in M$. The Matcher may first ask clarification questions and may later revise a selection. The final output is the ordered sequence $\hat{D}=(\hat{d}_1,\ldots,\hat{d}_k)$. Thus, the task adds clarification, cross-turn integration, and revision to one-shot matching under partial observability \cite{Clark1986}.

\begin{table*}[htb]
\centering
\small
\setlength{\tabcolsep}{6.75pt}
\renewcommand{\arraystretch}{1.08}
\begin{tabular}{llccccccc}
\toprule
\multirow{2}{*}{\textbf{Image Set}} 
& \multirow{2}{*}{\textbf{Source}} 
& \multicolumn{3}{c}{\textbf{Dataset Size}} 
& \multicolumn{4}{c}{\textbf{Context Properties}} \\
\cmidrule(lr){3-5} \cmidrule(lr){6-9}
& & \textbf{\# Items} & \textbf{\# Targets} & \textbf{\# Dialogues} 
& \textbf{Matching} & \textbf{Modality} & \textbf{Object Type} & \textbf{Nameable} \\
\midrule
Dogs      & W & 13 & 10 & 10 & Free-form & \textsc{Spoken} & \textsc{Real}     & \checkmark \\
Baskets 1 & W & 13 & 10 & 10 & Free-form & \textsc{Spoken} & \textsc{Real}     & -- \\
Baskets 2 & Z & 18 & 12 & 14 & Free-form & \textsc{Text}   & \textsc{Real}     & -- \\
Tangrams  & H & 12 & 12 & 14 & Sequential         & \textsc{Text}   & \textsc{Abstract} & -- \\
\bottomrule
\end{tabular}
\caption{
Dataset statistics and key properties of the four visual contexts used in our interactive visual grounding experiments. Source: W = \citet{wang-etal-2025-lvlms}; Z = \citet{zeng2026lvlmshumansgrounddifferently}; and H = \citet{Hawkins2020}. In the source human tasks, free-form matching permitted revisions, whereas sequential matching proceeded one target at a time without revision; our simulated interaction rules are described in Figure~\ref{fig:methodology} and Section~\ref{sec:interactive_vg}.
}
\label{tab:dataset_summary}
\end{table*}

\paragraph{Interaction Protocols}
We instantiate four protocols that vary how much target information is provided upfront and how much the Matcher must acquire through interaction.

In \staticprotocol{}, the Matcher receives one full human-produced target description at a time and selects without interaction. 
\fullprotocol{} provides the same description but permits optional follow-up questions, testing whether clarification can verify, refine, or repair an initially informative input. ``Full'' means that the description comes from a high-accuracy human dialogue and is therefore expected to be sufficiently informative, not that it exhaustively specifies the target. \partialprotocol{} instead begins with an underspecified attribute, requiring the Matcher to seek missing information. \oracleprotocol{} provides no initial description; the Director returns only yes/no/unclear answers, so the Matcher must decide what to request and construct a target representation through questions \cite{Vries2017CVPR}. Together, the protocols progress from matching an informative description (\staticprotocol{}) to optional clarification (\fullprotocol{}), recovery from underspecification (\partialprotocol{}), and fully question-driven grounding (\oracleprotocol{}).

A description drawn from a high-performing human dialogue may still leave an LVLM or even another human uncertain or with an incorrect prediction, motivating optional interaction even in the \fullprotocol{} condition.

\paragraph{Evaluation Metrics} 

We primarily consider \textbf{accuracy} and \textbf{question rate} as our main metrics, reflecting models' visual matching ability and tendency to seek information in interactive settings. Accuracy is the percentage of correctly matched items within a task round. Question rate is the average number of questions asked per item; by definition, it is 0 under \staticprotocol{}. As a baseline, random accuracy is $1/n$ for a Matcher view with $n$ candidates.

We also analyze \textbf{calibration} to assess whether models know when they have enough information to make a selection. Each Matcher is prompted to report a confidence score with its final selection, and we compare these stated confidence scores against empirical accuracy using calibration plots.\footnote{We also compute Expected Calibration Error (ECE), which is strongly correlated with accuracy in our setting (Spearman's $\rho=-0.96$), making it less informative as a separate measure of calibration. See Appendix~\ref{app:ece} for details.} 

Finally, we consider \textbf{self-repairs} as a secondary behavioral signal. Self-repairs occur when a Matcher revises an earlier selection after receiving additional information, indicating whether models can recognize and correct earlier grounding decisions during interaction.

\section{Experimental Setup \label{sec:experimental-setup}}

\subsection{Datasets}

We evaluate interactive visual grounding across four human-grounded, object-level visual contexts (Table~\ref{tab:dataset_summary}): dogs and baskets from \citet{wang-etal-2025-lvlms}, baskets from \citet{zeng2026lvlmshumansgrounddifferently}, and tangrams from \citet{Hawkins2020}. Appendix~\ref{app:datasets} provides more details on the four image sets, and Panel A of Figure~\ref{fig:methodology} shows representative examples.

We select these datasets for two reasons. First, they provide human-to-human reference data over the same visual objects, allowing us to simulate human Directors and compare model behavior against human-grounded baselines. Second, they cover several dimensions central to referential communication: real versus abstract objects, spoken versus text-based references, free matching versus sequential target selection, and objects with versus without conventional names. The less nameable contexts are particularly challenging: baskets require fine-grained attribute descriptions, while tangrams often elicit highly variable descriptions across speakers. We also include two basket image sets with different levels of initial-round human performance: 80\% in \citet{zeng2026lvlmshumansgrounddifferently} and 100\% in \citet{wang-etal-2025-lvlms}. Together, these image sets provide a diverse, human-grounded testbed for evaluating LVLMs under varying degrees of visual ambiguity, nameability, and referential difficulty.

All four datasets contain repeated-round human reference data, where a round denotes one pass through the same image set by the same dyad. By default, we use only the initial round, where referring expressions are more self-contained and less dependent on dyad-specific common ground.



\subsection{Referring Expression Processing}\label{sec:refexp}

We reuse the referring expressions from the four datasets in Table~\ref{tab:dataset_summary}, which contain human-produced descriptions for the same visual objects across multiple discourse participants. This supports a realistic and diverse evaluation setting, given the open-ended nature of referring expression generation.



\paragraph{Filtering} To ensure that the retained descriptions are accurate and sufficiently informative, we keep only initial-round referring expressions from dialogues in which human participants achieved at least 80\% matching accuracy for \citet{zeng2026lvlmshumansgrounddifferently} and 90\% for \citet{Hawkins2020}, balancing description quality against sample size. For the two image sets from \citet{wang-etal-2025-lvlms}, initial-round accuracy is 100\%, so we retain all dialogues. Although item-level success could provide a more fine-grained filter, we use dialogue-level accuracy because the referring expressions are drawn from complete interaction transcripts and may therefore depend on the broader dialogue context.

We further control description quality by selecting only participants whose matching performance does not decrease across rounds within each dataset. Since repeated reference games typically become easier as participants establish common ground, this filter removes dialogues with unstable engagement or inconsistent task behavior. It also ensures that the selected dialogues are suitable for our multi-round follow-up studies, where we examine whether LVLMs can handle realistic long-horizon interactive visual grounding. This filtering yields 48 high-quality human dialogues.

\paragraph{Full Target Item Description Extraction}
Following \citet{zeng2026lvlmshumansgrounddifferently}, we use GPT-5.4~\cite{openai_gpt5_2025} to automatically extract object-level referring expressions for each image set in Table~\ref{tab:dataset_summary}, except for Tangrams, which already includes manually curated object-level descriptions. This process yields 536 diverse, high-quality referring expressions across the four image sets, which we use in \staticprotocol{} and \fullprotocol{}.

\begin{figure*}[]
    \centering    \includegraphics[width=1\linewidth]{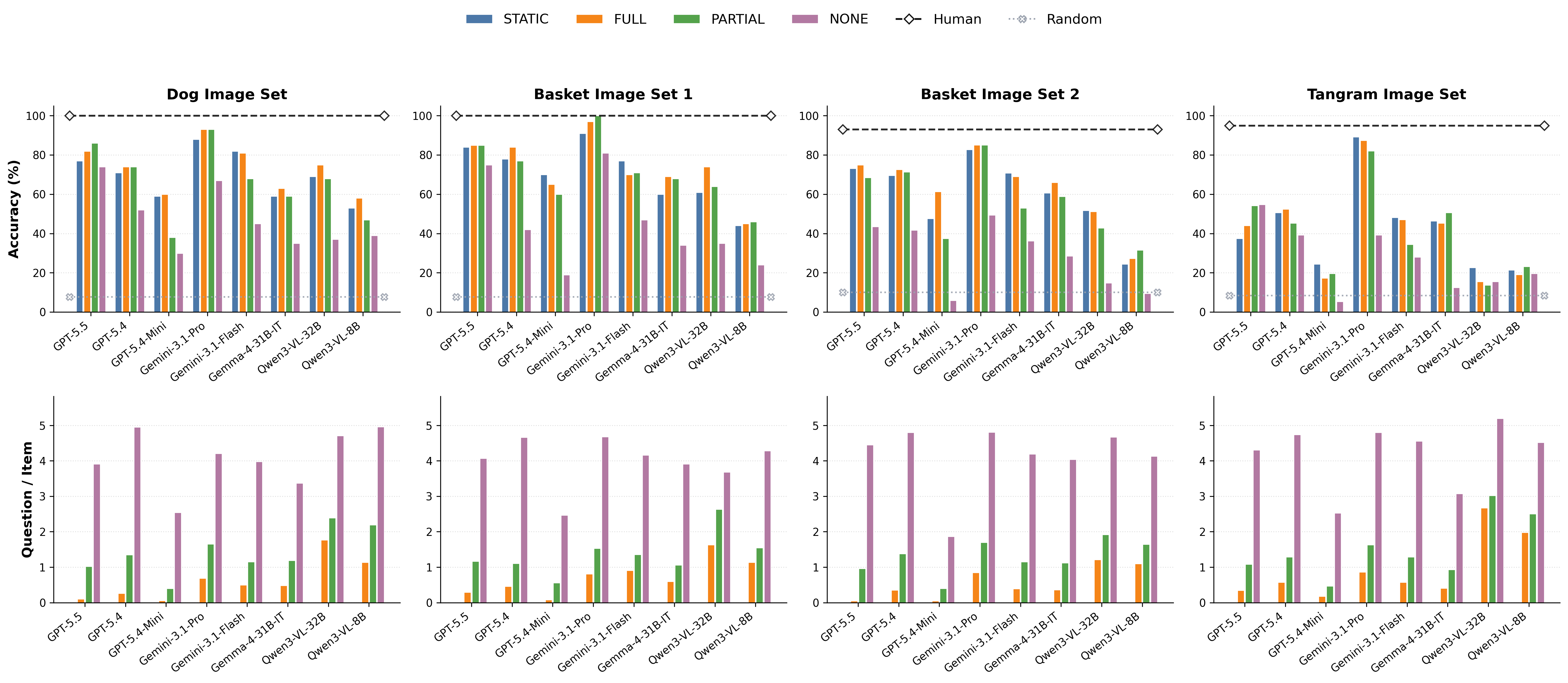}
    \caption{Overall results across the four image sets. Top: average matcher accuracy. 
    Bottom: average questions per target item. Colors denote the four interaction protocols; horizontal lines mark human and random baselines. The human line is a \textit{task-level} reference, but not a protocol-matched baseline, since the interaction conditions differ.}
    \label{fig:overallResults}
\end{figure*}

\paragraph{Partial Target Item Description Extraction}
To construct underspecified referring expressions, we prompt GPT-5.4 with the full descriptions for all target items in the same image set and ask it to extract the least informative attribute from each description. We use these extracted attributes as the initial descriptions in \partialprotocol{}. This LLM-as-a-judge procedure is designed not to find optimal shortened descriptions, but to minimize initial informativeness so that the Matcher must decide whether clarification is needed. We verify that the extracted attributes are lexically supported by the source full expressions (see Appendix~\ref{app:partial_description_validation}). Our experiments further validate this design: all models ask questions under \partialprotocol{}, indicating that the extracted attributes are underspecified enough to elicit clarification (e.g., Figure~\ref{fig:overallResults} in Section~\ref{sec:results}).

\subsection{Director Simulation}

Because our focus is the Matcher, we use GPT-5.4 as a fixed simulated Director. This choice is motivated by prior evidence that GPT-5.2 reaches about 85\% initial-round accuracy with human Matchers \cite{zeng2026lvlmshumansgrounddifferently}. Fixing the Director makes information access more comparable across Matchers and protocols; Section~\ref{sec:follw-up-studies} tests sensitivity to an alternative Director, showing comparable results.

The Director is always conditioned on the same human-produced full target description, defined in Section~\ref{sec:refexp}. In \staticprotocol{}, we programmatically present this preconfigured full description to the Matcher one target at a time. In \fullprotocol{} and \partialprotocol{}, the Matcher first receives a preconfigured full or underspecified description, respectively, and the GPT-5.4 Director is called only when the Matcher asks a clarification question. In \oracleprotocol{}, no initial description is provided; instead, the Director answers the Matcher's questions while being prompted with the corresponding full target description. When a question asks about information not explicitly stated in the description, the prompt instructs the Director to answer based only on the provided description and visual input. See Appendix~\ref{app:prompts} for the full Director prompts.

\subsection{Matcher Simulation}

\paragraph{Models}
We evaluate LVLMs from different model families and scales to simulate matchers: (1) five proprietary LVLMs, namely GPT-5.4, GPT-5.4-Mini, and GPT-5.5 \cite{openai2026gpt54,openai2026gpt55}, and Gemini-3.1-Pro and Gemini-3.1-Flash-Lite \cite{googledeepmind2026gemini31pro}; and (2) three open-weight LVLMs, namely Gemma-4-31B-IT \cite{google2026gemma4} and Qwen3-VL-8B/32B-Instruct \cite{bai2025qwen3vl}.

\paragraph{Prompting} We use protocol-specific system prompts while keeping the task overview fixed for both the Director and Matcher across conditions. In interactive settings, the Matcher returns a JSON object containing a natural-language utterance and any selections it makes, with a 100-point confidence score for each selection. In the static setting, the Matcher returns a JSON selection with a confidence score. See Appendix~\ref{app:prompts} for full prompts.

\paragraph{Simulation Scale}
Across the four datasets, we obtain 536 item--description pairs. Under the four interaction protocols in the main experiments, each LVLM matcher makes 2,144 target selections based on information from the simulated Director. Across eight LVLM matchers, this yields 17,152 selections. Including the follow-up studies in Section~\ref{sec:follw-up-studies}, which add another 21,696 selections, our experiments produce 38,848 total target selections for analysis.

\subsection{Multi-Turn Dialogue Simulation}

\paragraph{Visual Input Format}
For each image set, the Director is shown only the target items, while the Matcher is shown the full candidate set, which may include distractors. We concatenate the individual item images into a single indexed grid and provide the grid as one image input. Director indices follow the item-description order in the original dataset, while Matcher indices are randomized to prevent trivial index matching or data contamination.

\paragraph{Turn Taking}

The Director communicates first in \staticprotocol{}, \fullprotocol{}, and \partialprotocol{}, whereas the Matcher initiates \oracleprotocol{} with a yes/no question. In all three interactive protocols, namely, \fullprotocol{}, \partialprotocol{}, and \oracleprotocol{}, Matchers may ask questions and revise earlier selections when later information changes their decision. Each turn contains one natural-language utterance capped at 50 words, and interactions are limited to a 10-turn budget. If this budget is exhausted, the Matcher is prompted to make a forced target selection to ensure termination. The Director advances to the next target only after the Matcher selects. The final prediction sequence is computed from the Matcher’s finalized selections.


\begin{table}[t]
\centering
\resizebox{\linewidth}{!}{%
\begin{tabular}{llllll}
\toprule
Model & Dogs & Baskets 1 & Baskets 2 & Tangrams & All \\
\midrule
GPT-5.5 & \cellcolor{red!11}$-14.0^{\tiny **}$ & \cellcolor{red!11}$-15.0^{\tiny **}$ & \cellcolor{red!13}$-18.5^{\tiny ***}$ & \cellcolor{red!24}$-40.5^{\tiny ***}$ & \cellcolor{red!17}$-25.3^{\tiny ***}$ \\
GPT-5.4 & \cellcolor{red!17}$-26.0^{\tiny ***}$ & \cellcolor{red!12}$-16.0^{\tiny ***}$ & \cellcolor{red!14}$-20.8^{\tiny ***}$ & \cellcolor{red!25}$-42.9^{\tiny ***}$ & \cellcolor{red!18}$-27.3^{\tiny ***}$ \\
GPT-5.4-Mini & \cellcolor{red!24}$-40.0^{\tiny ***}$ & \cellcolor{red!19}$-30.0^{\tiny ***}$ & \cellcolor{red!20}$-32.1^{\tiny ***}$ & \cellcolor{red!39}$-70.8^{\tiny ***}$ & \cellcolor{red!28}$-47.7^{\tiny ***}$ \\
Gemini-3.1-Pro & \cellcolor{red!7}$\mathbf{-7.0^{\tiny *}}$ & $\mathbf{0.0}$ & \cellcolor{red!8}$\mathbf{-8.3^{\tiny *}}$ & \cellcolor{red!7}$\mathbf{-6.0}$ & \cellcolor{red!7}$\mathbf{-6.8^{\tiny ***}}$ \\
Gemini-3.1-Flash & \cellcolor{red!13}$-18.0^{\tiny ***}$ & \cellcolor{red!15}$-23.0^{\tiny **}$ & \cellcolor{red!15}$-22.6^{\tiny ***}$ & \cellcolor{red!27}$-47.0^{\tiny ***}$ & \cellcolor{red!18}$-28.9^{\tiny ***}$ \\
Gemma-4-31B-IT & \cellcolor{red!22}$-37.0^{\tiny ***}$ & \cellcolor{red!19}$-31.0^{\tiny ***}$ & \cellcolor{red!18}$-27.4^{\tiny ***}$ & \cellcolor{red!26}$-44.6^{\tiny ***}$ & \cellcolor{red!22}$-36.7^{\tiny ***}$ \\
Qwen3-VL-32B & \cellcolor{red!16}$-25.0^{\tiny ***}$ & \cellcolor{red!17}$-26.0^{\tiny **}$ & \cellcolor{red!25}$-41.7^{\tiny ***}$ & \cellcolor{red!40}$-72.6^{\tiny ***}$ & \cellcolor{red!27}$-46.2^{\tiny ***}$ \\
Qwen3-VL-8B & \cellcolor{red!25}$-42.0^{\tiny ***}$ & \cellcolor{red!31}$-54.0^{\tiny ***}$ & \cellcolor{red!35}$-61.9^{\tiny ***}$ & \cellcolor{red!40}$-72.0^{\tiny ***}$ & \cellcolor{red!34}$-61.4^{\tiny ***}$ \\
\bottomrule
\end{tabular}
}
\caption{Best paired mean LVLM--human matcher accuracy gap (\%) across datasets, selected from the best-performing interaction protocol for each model--dataset pair. The All column is computed over all examples from the four datasets. Darker red denotes larger deficits relative to human matchers, and bold marks the smallest gap within each dataset. Significance is assessed with paired $t$-tests: $^{\tiny *}p \le .05$, $^{\tiny **}p \le .01$, and $^{\tiny ***}p \le .001$. See Appendix~\ref{app:lvlm_human_gap} for protocol-specific results.}
\label{tab:lvlm_human_gap_heatmap}
\end{table}

\begin{table}[t]
\centering
\resizebox{\linewidth}{!}{%
\begin{tabular}{lllll}
\toprule
Model & \staticprotocol{} & \fullprotocol{} & \partialprotocol{} & \oracleprotocol{} \\
\midrule
GPT-5.5 & $65.8$ & $\underline{69.5}_{\scriptscriptstyle \textcolor{ForestGreen}{+3.6}}$ & $\underline{71.4}_{\scriptscriptstyle \textcolor{ForestGreen}{+5.5}}$ & $\mathbf{59.7}_{\scriptscriptstyle \textcolor{red}{-6.1}}$ \\
GPT-5.4 & $66.1$ & $69.4_{\scriptscriptstyle \textcolor{ForestGreen}{+3.4*}}$ & $65.5_{\scriptscriptstyle \textcolor{red}{-0.4}}$ & $43.2_{\scriptscriptstyle \textcolor{red}{-23.6*}}$ \\
GPT-5.4-Mini & $47.9$ & $49.0_{\scriptscriptstyle \textcolor{ForestGreen}{+0.6}}$ & $37.1_{\scriptscriptstyle \textcolor{red}{-11.5*}}$ & $13.5_{\scriptscriptstyle \textcolor{red}{-35.2*}}$ \\
Gemini-3.1-Pro & $\mathbf{87.5}$ & $\mathbf{89.9}_{\scriptscriptstyle \textcolor{ForestGreen}{+2.9}}$ & $\mathbf{89.0}_{\scriptscriptstyle \textcolor{ForestGreen}{+2.3}}$ & $\underline{56.7}_{\scriptscriptstyle \textcolor{red}{-28.6*}}$ \\
Gemini-3.1-Flash & $\underline{67.8}$ & $65.3_{\scriptscriptstyle \textcolor{red}{-2.7}}$ & $54.5_{\scriptscriptstyle \textcolor{red}{-12.9*}}$ & $37.9_{\scriptscriptstyle \textcolor{red}{-30.4**}}$ \\
Gemma-4-31B-IT & $56.0$ & $60.0_{\scriptscriptstyle \textcolor{ForestGreen}{+4.3}}$ & $58.4_{\scriptscriptstyle \textcolor{ForestGreen}{+2.6}}$ & $26.4_{\scriptscriptstyle \textcolor{red}{-29.0**}}$ \\
Qwen3-VL-32B & $48.8$ & $50.5_{\scriptscriptstyle \textcolor{ForestGreen}{+2.8}}$ & $44.0_{\scriptscriptstyle \textcolor{red}{-4.0}}$ & $23.9_{\scriptscriptstyle \textcolor{red}{-25.5*}}$ \\
Qwen3-VL-8B & $33.6$ & $35.0_{\scriptscriptstyle \textcolor{ForestGreen}{+1.6}}$ & $35.3_{\scriptscriptstyle \textcolor{ForestGreen}{+1.2}}$ & $21.6_{\scriptscriptstyle \textcolor{red}{-12.7*}}$ \\
\bottomrule
\end{tabular}
}
\caption{Micro-averaged accuracy (\%) by model and interaction protocol. Subscripts report paired mean differences relative to \staticprotocol{}; \textcolor{ForestGreen}{green} indicates gains and \textcolor{red}{red} indicates losses. The best-performing model within each protocol is \textbf{bolded}, and the second-best model is \underline{underlined}. Significance is assessed with paired $t$-tests: $^{*}p \le .05$, $^{**}p \le .01$, and $^{***}p \le .001$.}
\label{tab:protocol_aggregate_accuracy}
\end{table}

\section{Results\label{sec:results}}

\subsection{Interactive Visual Grounding Poses a Challenging Task}

\paragraph{LVLM matchers significantly lag behind human matchers in interactive visual grounding.}
The upper panels of Figure~\ref{fig:overallResults} show that nearly all tested LVLM matchers underperform the corresponding task-level human baselines across the four image sets, with the largest gaps on the Tangram Image Set. Table~\ref{tab:lvlm_human_gap_heatmap} further shows that these gaps are often substantial and statistically significant under paired $t$-tests, especially when results are pooled across all datasets. The only case of human parity is Gemini-3.1-Pro on Basket Image Set 2 under \partialprotocol{}, but no model outperforms the human baselines. Moreover, the consistent gaps under \staticprotocol{} and \fullprotocol{} suggest that current LVLMs still struggle to comprehend descriptions
drawn from high-accuracy human dialogues \cite{tang-etal-2024-grounding, wang-etal-2025-lvlms, zeng2026lvlmshumansgrounddifferently}.

\paragraph{The task is challenging because it requires both visual matching and effective information seeking.}
GPT-5.4-Mini consistently asks the fewest questions. It outperforms Qwen3-VL-8B under \staticprotocol{} and \fullprotocol{}, but its advantage nearly disappears under \partialprotocol{} and reverses under \oracleprotocol{} (Table~\ref{tab:protocol_aggregate_accuracy}), where question asking matters more. This suggests that limited question asking can hurt when the task requires acquiring missing information. However, asking more questions alone is not sufficient. Gemini-3.1-Pro asks more questions than GPT-5.5, yet GPT-5.5 achieves higher overall accuracy under \oracleprotocol{}. These contrasts suggest that success requires not only visual matching or more questions, but useful questions and effective integration of answers across turns.


\subsection{Stronger Models Perform Better, but Question-Only Grounding Remains Hard}

\paragraph{Model performance exhibits a clear hierarchy.}
Figure~\ref{fig:overallResults} and Table~\ref{tab:protocol_aggregate_accuracy} show that larger models generally outperform their smaller counterparts from the same model family, suggesting benefits from scaling, and proprietary models tend to outperform open-weight models. Gemini-3.1-Pro is strongest overall, achieving the best performance under \staticprotocol{}, \fullprotocol{}, and \partialprotocol{}, with especially large gains on the Tangram Image Set. GPT-5.5 is also competitive, particularly under \oracleprotocol{}, suggesting stronger question-driven information seeking and integration capabilities.

\paragraph{LVLMs can benefit from interaction, but question-driven grounding remains challenging.}
Table~\ref{tab:protocol_aggregate_accuracy} shows that \fullprotocol{} improves over \staticprotocol{} for seven of the eight models, suggesting that optional follow-up questions can help refine or repair initial descriptions. Under \partialprotocol{}, four models improve over \staticprotocol{} despite receiving an underspecified description, consistent with successful information recovery, while four decline. Compared with \fullprotocol{}, however, six of the eight models perform worse under \partialprotocol{}, indicating that interaction often does not fully compensate for missing initial information.
The clearest limitation appears under \oracleprotocol{}, where all models perform below \staticprotocol{} despite asking the most questions. This suggests that current LVLMs struggle to construct and ground target representations through proactive question-driven dialogue.

\begin{figure}[]
    \centering    \includegraphics[width=1\linewidth]{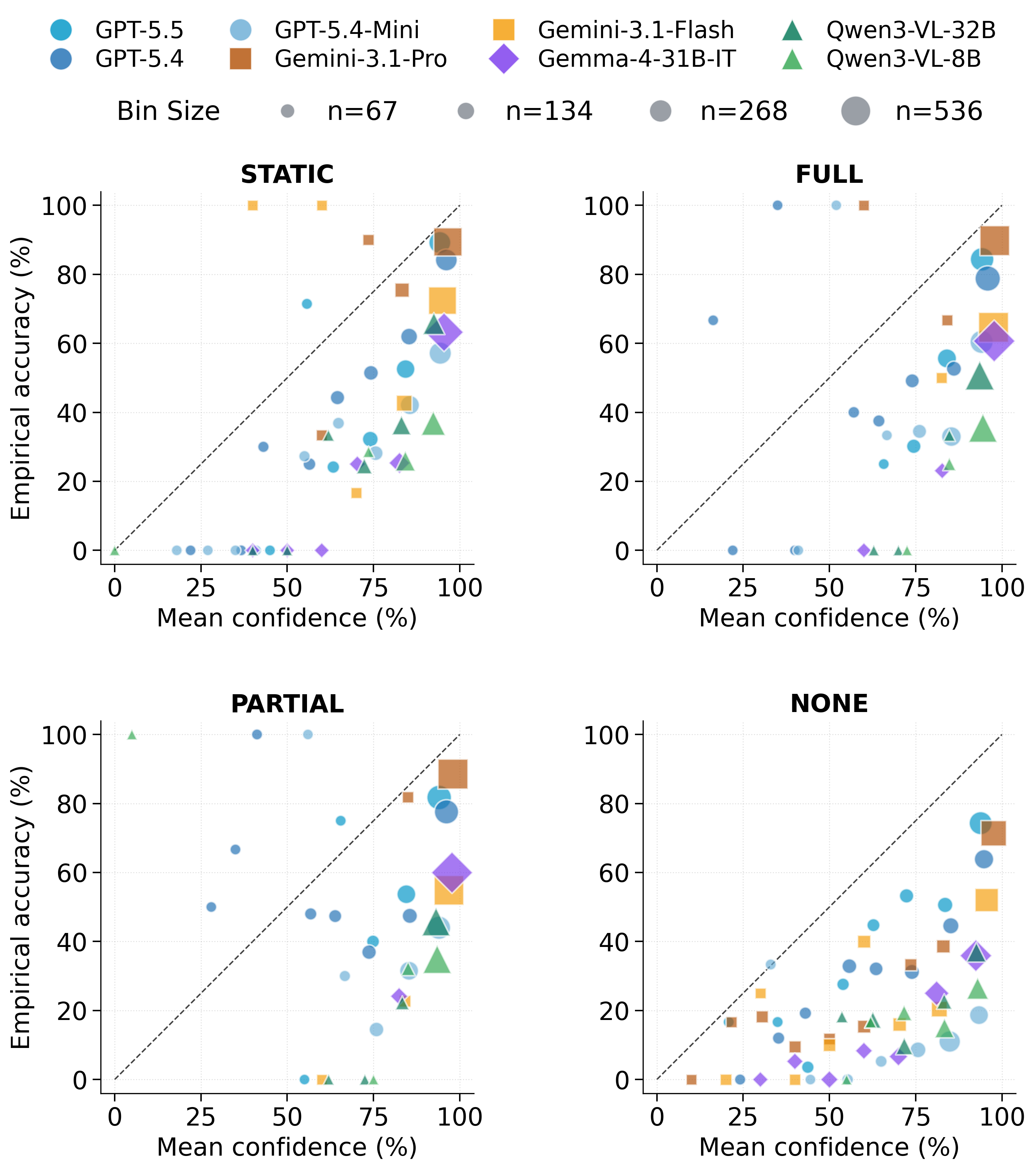}
    \caption{Protocol-level calibration. Points show model-specific confidence bins from 10 evenly spaced bins between 0 and 100; point size is proportional to bin count. The dashed diagonal denotes perfect calibration.}
    \label{fig:protocol_level_confidence_calibration}
\end{figure}

\subsection{LVLMs Are Not Yet Calibrated for Interactive Visual Grounding}
\label{sec:results_calibration_behavior}

\paragraph{Current LVLMs are sensitive to information availability, but remain overconfident.}
All models ask more questions when less initial information is available: \oracleprotocol{} $>$ \partialprotocol{} $>$ \fullprotocol{} $>$ \staticprotocol{}
($=0$ by design; lower panels of Figure~\ref{fig:overallResults}). Figure~\ref{fig:protocol_level_confidence_calibration} shows a similar shift in confidence: models tend to report lower confidence in more interaction-heavy protocols. However, most points still fall below the diagonal, indicating that reported confidence often exceeds empirical accuracy. This overconfidence is especially pronounced under \partialprotocol{} and \oracleprotocol{}, suggesting that LVLMs are less reliable at judging information sufficiency when the initial referring expression is underspecified or must be constructed through interaction. Overall, this suggests that LVLMs adjust their behavior to the amount of available information, but their confidence does not yet reliably reflect whether they have enough evidence for accurate grounding.


\paragraph{Self-repairs are rare and inconsistently reliable.}
Revising a selection could indicate that a model recognizes and acts on uncertainty, but repairs occur only under \fullprotocol{} and \partialprotocol{} and account for under 2\% of matches. Among the 49 observed repairs, 26 (53.1\%) correct a wrong selection, 5 (10.2\%) change a correct selection to a wrong one, and 18 (36.7\%) replace one wrong answer with another. Outcomes are strongly model-dependent: Gemini-3.1-Pro corrects 13 of its 17 revisions (76.5\%), whereas GPT-5.4 corrects none of six and nevertheless raises its confidence by 11.2 points on average. Thus, models can occasionally exploit later evidence to repair an error, but revision remains too sparse and its confidence changes too inconsistent to serve as reliable error awareness. Full model- and protocol-level results appear in Appendix~\ref{app:self-repair}.

\subsection{Summary}

Overall, interactive visual grounding remains challenging for current LVLMs. Models ask more questions when information is limited and can sometimes benefit from interaction, especially when follow-up questions refine or repair an initial description. However, they still lag behind human matchers, struggle with proactive question-driven dialogue, and remain poorly calibrated for the task.

\section{Follow-Up Studies\label{sec:follw-up-studies}}

\begin{figure}[]
    \centering    \includegraphics[width=\linewidth]{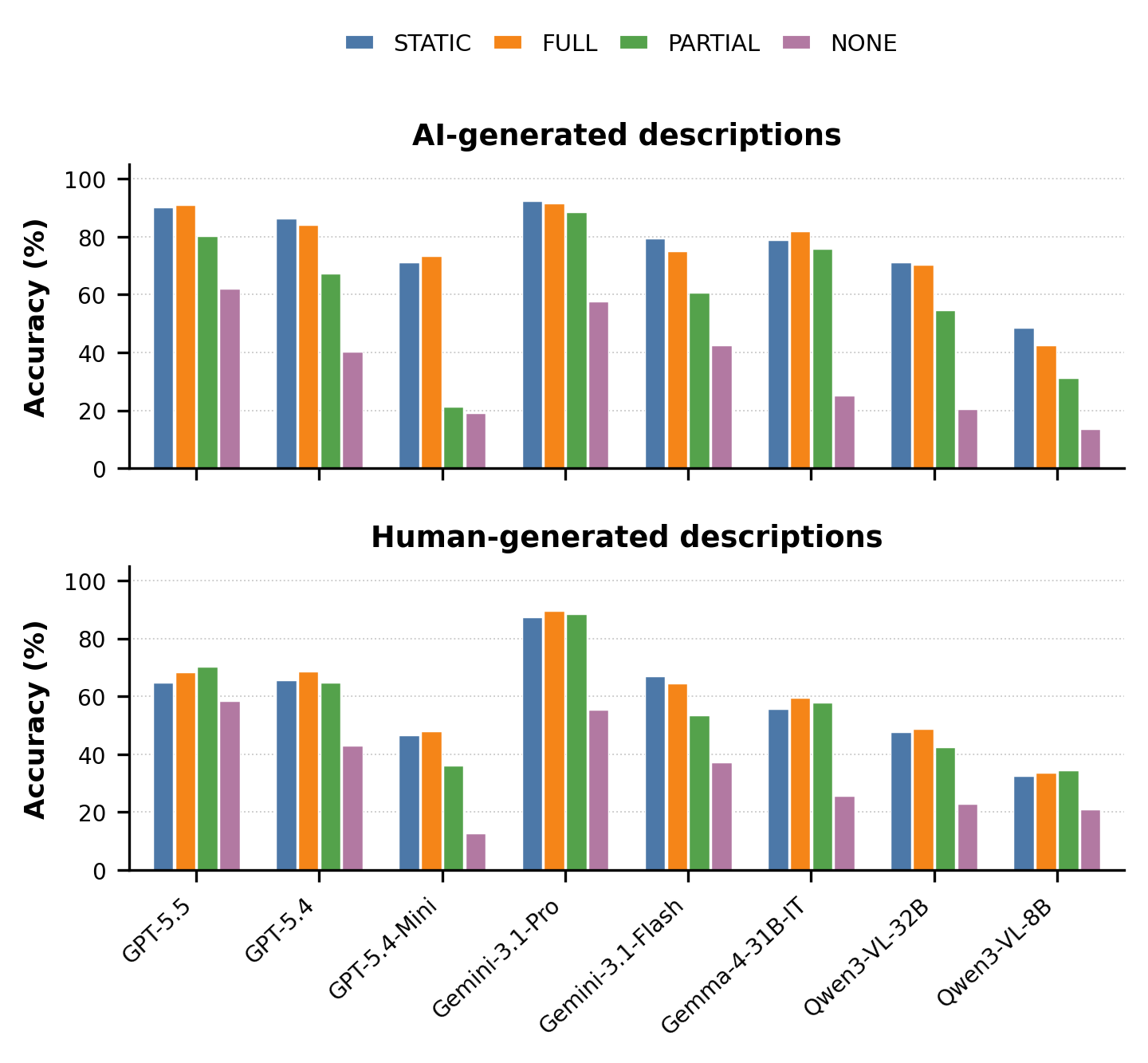}
    \caption{Accuracy (\%) on AI-generated and human-generated descriptions across models and interaction protocols, averaged over all four datasets.}\label{fig:overall_human_versus_ai_accuracy}
\end{figure}

\begin{table}[t]
\centering
\small
\begin{tabular}{lllll}
\toprule
\multirow{2}{*}{Reasoning} 
& \multicolumn{2}{c}{GPT-5.4} 
& \multicolumn{2}{c}{GPT-5.4-Mini} \\
\cmidrule(lr){2-3} \cmidrule(lr){4-5}
& \staticprotocol{} & \fullprotocol{} & \staticprotocol{} & \fullprotocol{} \\
\midrule
None   
& $50.6$ & $52.4$ & $24.4$ & $17.3$ \\
Medium 
& $52.4_{\scriptscriptstyle \textcolor{ForestGreen}{+1.8}}$
& $60.7_{\scriptscriptstyle \textcolor{ForestGreen}{+8.3}}$
& $26.2_{\scriptscriptstyle \textcolor{ForestGreen}{+1.8}}$
& $30.4_{\scriptscriptstyle \textcolor{ForestGreen}{+13.1}}$ \\
High   
& $48.8_{\scriptscriptstyle \textcolor{red}{-1.8}}$
& $61.3_{\scriptscriptstyle \textcolor{ForestGreen}{+8.9}}$
& $15.5_{\scriptscriptstyle \textcolor{red}{-8.9}}$
& $31.0_{\scriptscriptstyle \textcolor{ForestGreen}{+13.7}}$ \\
\bottomrule
\end{tabular}
\caption{Accuracy (\%) under different reasoning-effort settings on tangrams. Subscripts report differences relative to default no reasoning.}
\label{tab:reasoning_effort_accuracy}
\end{table}

To test whether our conclusions depend on description source, test-time reasoning, repeated-reference structure, Director choice, or the original visual contexts, we conduct five follow-up studies. We (i) replace human descriptions with GPT-5.4-generated descriptions across all Matchers and protocols; (ii) vary reasoning effort for GPT-5.4 and GPT-5.4-Mini on tangrams; (iii) evaluate three LVLMs over four rounds of repeated reference, using the corresponding human-produced expressions from the original human--human dialogues; (iv) replace GPT-5.4 with Gemini-3.1-Flash as the Director on Basket Set 2 and Tangrams; and (v) evaluate five LVLM Matchers (i.e., GPT-5.4, GPT-5.4-Mini, Gemini-3.1-Flash, Gemma-4-31B-IT, and Qwen3-VL-8B) on six public face image sets with GPT-5.4-generated descriptions \cite{Holland2018}, each containing 12 images of one individual with varied facial expressions.

\paragraph{LVLMs perform better with AI-generated descriptions, but result patterns remain comparable.}
Figure~\ref{fig:overall_human_versus_ai_accuracy} shows that LVLMs generally achieve higher accuracy with AI-generated descriptions than with human-generated descriptions, especially under \staticprotocol{} and \fullprotocol{}. This suggests that AI-generated descriptions are easier for LVLMs to interpret, while human referring expressions pose a distinct challenge. Importantly, however, the overall protocol-level patterns remain similar: \staticprotocol{} and \fullprotocol{} are strongest, \partialprotocol{} is more variable, and \oracleprotocol{} remains most difficult. 

\begin{figure}[]
    \centering    \includegraphics[width=1.0\linewidth]{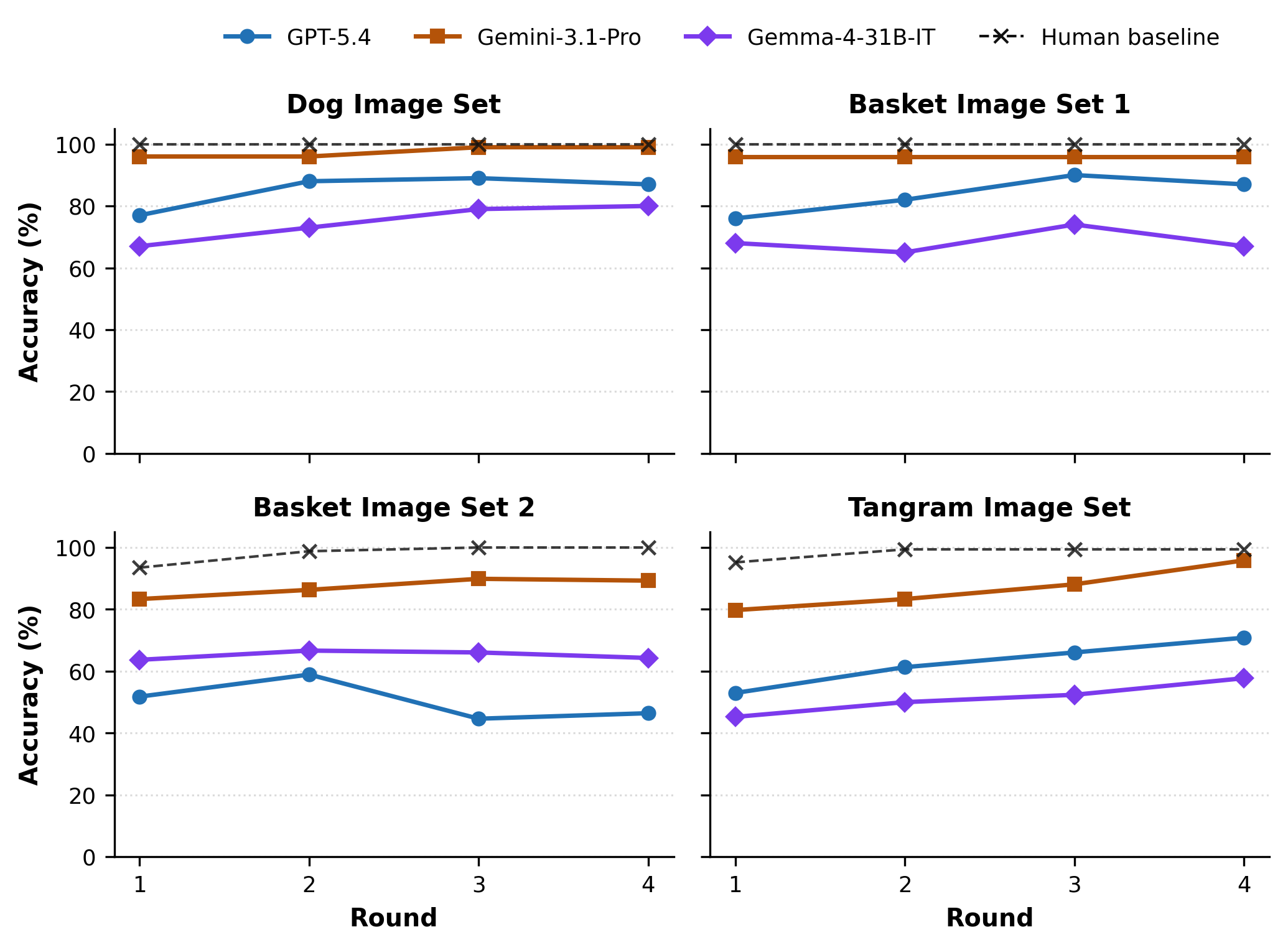}
    \caption{Accuracy (\%) across rounds on all four image sets under \fullprotocol{}. See Appendix~\ref{app:multi-round} for \staticprotocol{} results.}\label{fig:full_multi_round_accuracy_by_dataset_2x2}
\end{figure}

\paragraph{Reasoning helps in interactive settings, but scaling plateaus quickly.}
Reasoning yields little benefit in \staticprotocol{}, but substantially improves accuracy in \fullprotocol{} (Table~\ref{tab:reasoning_effort_accuracy}). This suggests that stronger reasoning helps models gather and synthesize visual evidence through conversation. However, gains largely plateau from medium to high reasoning, indicating that additional test-time compute alone is not sufficient. Reasoning also modestly improves calibration, with similar plateaus from medium to high effort, as discussed in Appendix~\ref{app:reasoning_effort}.

\paragraph{LVLMs can benefit from repeated interaction, but the gap to human matchers remains.}
Figure~\ref{fig:full_multi_round_accuracy_by_dataset_2x2} shows that models can improve across rounds under \fullprotocol{}. Results for \staticprotocol{} show similar patterns and are reported in Appendix~\ref{app:multi-round}. These findings suggest that LVLMs can benefit from increasingly efficient referring expressions produced through entrainment between humans. However, improvements are inconsistent across datasets, and human matchers continue to outperform LVLMs. While Gemini-3.1-Pro approaches human baselines in some settings, long-horizon interactive visual grounding remains a broad challenge for LVLMs.


\begin{table}[t]
\centering
\small
\setlength{\tabcolsep}{3pt}
\begin{tabular}{lrr}
\toprule
\multicolumn{3}{l}{\textbf{A. Alternative Director robustness}} \\
Director / comparison & \fullprotocol{} & \partialprotocol{} \\
\midrule
GPT-5.4 & 51.5 & 48.3 \\
Gemini-3.1-Flash & 50.4 & 44.8 \\
$\Delta$ (Gemini $-$ GPT) & $-1.1$ & $-3.5$ \\
Rank correlation $\rho$ & $.929$ & $.976$ \\
\bottomrule
\end{tabular}

\vspace{2pt}
\begin{tabular}{lrrrr}
\toprule
\multicolumn{5}{l}{\textbf{B. Face-domain robustness}} \\
Metric & \staticprotocol{} & \fullprotocol{} & \partialprotocol{} & \oracleprotocol{} \\
\midrule
Accuracy (\%) & 48.7 & 47.8 & 41.9 & 30.7 \\
$\Delta$ vs. \staticprotocol{} & -- & $-0.9$ & $-6.8$ & $-18.0$ \\
\bottomrule
\end{tabular}
\caption{Aggregate alternative-Director (A) and face-domain (B) results. Panel A averages across Matchers, datasets, and description sources; Panel B across Matchers and six face sets.}
\label{tab:additional_robustness_results}
\end{table}

\paragraph{Matcher rankings are stable across Director models.} Table~\ref{tab:additional_robustness_results} A shows that replacing the GPT-5.4 Director with Gemini-3.1-Flash reduces mean Matcher accuracy by only 1.1 points in \fullprotocol{} (51.5\% to 50.4\%), but by 3.5 points in \partialprotocol{} (48.3\% to 44.8\%). The larger \partialprotocol{} decrease is consistent with the Director carrying more responsibility when the initial description is underspecified. Despite these absolute changes, Matcher rankings remain highly correlated across Directors (Spearman's $\rho=.929$ for \fullprotocol{} and $.976$ for \partialprotocol{}). Thus, interaction quality affects accuracy, especially under underspecification, but the central Matcher hierarchy is not an artifact of a single simulated Director (Appendix~\ref{app:alternative-director}).

\paragraph{The protocol-level pattern extends to fine-grained face matching.} Across six face sets with AI-generated descriptions, Table~\ref{tab:additional_robustness_results}B shows that \fullprotocol{} remains within 0.9 points of \staticprotocol{} (47.8\% vs. 48.7\%), whereas \partialprotocol{} and \oracleprotocol{} fall 6.8 and 18.0 points below \staticprotocol{}, respectively. Thus, optional clarification largely preserves performance when an informative description is available, but interaction does not fully compensate as initial information is removed. This graded pattern matches Figure~\ref{fig:overall_human_versus_ai_accuracy} and extends the main protocol-level observation to a distinct fine-grained domain, confirming the core finding in Section~\ref{sec:results} (Appendix~\ref{app:face-domain}).

\section{Conclusion}

We introduce a controlled framework for evaluating interactive visual grounding beyond one-shot matching. Across four human-grounded visual contexts, current LVLMs perform significantly below task-level human references. Interaction can help when follow-up questions refine or repair an initial description, but models consistently struggle with proactive question-driven grounding. They are also poorly calibrated, often reporting confidence that exceeds accuracy. These findings are further validated across follow-up studies using AI-generated descriptions, test-time scaling, exposure to repeated interaction, an alternative simulated Director, and a fine-grained face context. Overall, interactive visual grounding remains an important challenge for LVLMs, with potential implications for future embodied human--AI interaction.

\section*{Acknowledgments}

This material is based upon work supported by the National Science Foundation (NSF) under Grant No.~2125295 and by a seed grant from Stony Brook University. Any opinions, findings, and conclusions or recommendations expressed in this material are those of the author(s) and do not necessarily reflect the views of the NSF.

We are grateful for support from the Institute for Advanced Computational Science (IACS) at Stony Brook University. We also thank the three anonymous reviewers for their valuable feedback.

\section*{AI Usage Disclosure}

We used generative AI tools, particularly Codex, to assist with developing experimental and analysis code and with language polishing. All methodological and interpretive decisions were made by the authors, who reviewed and verified all AI-assisted outputs.

\section*{Limitations}

\paragraph{Simulated Directors and Director Dependence.} We use simulated rather than human Directors for scale and experimental control. Repeating \fullprotocol{} and \partialprotocol{} with Gemini-3.1-Flash produces highly stable Matcher rankings, reducing concern that the main conclusions are specific to GPT-5.4. That said, this robustness check does not establish equivalence to human Directors, and individual model--protocol outcomes can still vary with the Director.

\paragraph{Attribution.} The four protocols vary how much information must be acquired interactively. They are designed to create a controlled way to separate visual matching from information seeking and question-driven grounding, but do not uniquely isolate question selection, answer integration, dialogue memory, and visual matching etc. Our claim is also limited: these protocols increase the need for information seeking and synthesis beyond static visual matching, rather than attributing each failure to one precise subcomponent, which we leave to future studies.

\paragraph{Visual Context Coverage.}
The main testbed contains four object-level contexts, and the face follow-up adds another fine-grained within-category setting. Neither covers cluttered scenes, spatial or relational grounding, or open-world visual search. Our conclusions therefore concern proactive interactive grounding in controlled, fine-grained matching contexts rather than visual grounding in all settings.

\paragraph{Constrained Interaction under {\oracleprotocol{}}.} The Director provides yes/no answers under \oracleprotocol{} to create a controlled information-seeking setting, following \citet{Vries2017CVPR}. Open-ended Director answers may support different questioning strategies and should be studied separately.

\section*{Ethical Considerations}

\paragraph{Data Use and Privacy.}
This work evaluates LVLMs on interactive visual grounding using existing research datasets and public or previously released visual materials. We do not collect new human-subject data. The face-domain robustness study uses public research images only for within-set visual matching; we neither infer identity nor predict sensitive attributes. The main evaluated images involve object-level visual contexts such as dogs, baskets, and abstract tangrams, which reduces risks related to privacy or demographic stereotyping.

\bibliography{custom}

\appendix

\startcontents[appendix]
\section*{Appendix Table of Contents}
\printcontents[appendix]{}{1}{\setcounter{tocdepth}{2}}

\begin{table*}[t]
\centering
\resizebox{\linewidth}{!}{%
\begin{tabular}{llll}
\toprule
Model & Provider / Family & Variant used (snapshot / HF repo) & Inference backend \\
\midrule
GPT-5.5 \cite{openai2026gpt55} & OpenAI & \texttt{gpt-5.5-2026-04-23} & OpenAI API \\
GPT-5.4 \cite{openai2026gpt54} & OpenAI & \texttt{gpt-5.4-2026-03-05} & OpenAI API \\
GPT-5.4-Mini \cite{openai2026gpt54} & OpenAI & \texttt{gpt-5.4-mini-2026-03-17} & OpenAI API \\
\midrule
Gemini-3.1-Pro \cite{googledeepmind2026gemini31pro} & Google Gemini & \texttt{gemini-3.1-pro-preview} & Gemini API \\
Gemini-3.1-Flash \cite{googledeepmind2026gemini31pro} & Google Gemini & \texttt{gemini-3.1-flash-lite-preview} & Gemini API \\
\midrule
Gemma-4-31B-IT \cite{google2026gemma4} & Gemma & \texttt{google/gemma-4-31b-it} & vLLM \\
Qwen3-VL-32B \cite{bai2025qwen3vl} & Qwen & \texttt{Qwen/Qwen3-VL-32B-Instruct} & vLLM \\
Qwen3-VL-8B \cite{bai2025qwen3vl} & Qwen & \texttt{Qwen/Qwen3-VL-8B-Instruct} & vLLM \\
\bottomrule
\end{tabular}
}
\caption{LVLM variants used in our experiments. API models are identified by their provider-facing model names, and open-weight models are identified by their Hugging Face repository IDs. Open-weight models are run locally with vLLM.}
\label{tab:model_variants}
\end{table*}

\section{LVLM Details\label{app:lvlms}}

Table~\ref{tab:model_variants} lists the exact LVLM variants used in our experiments. We evaluate proprietary GPT and Gemini models through their respective APIs, using the latest available variants at the time of experimentation. For open-weight models, we use the Hugging Face checkpoints listed in the table and run local inference with vLLM \cite{Kwon2023}. Unless otherwise specified, we use the default decoding configuration across models.



\section{Expected Calibration Error\label{app:ece}}

We compute Expected Calibration Error (ECE) to summarize the alignment between the Matcher's stated confidence and empirical accuracy. For each final selection, the Matcher is prompted to report a confidence score $c_j \in [0,100]$, and we define correctness as whether the selected item matches the target item at the corresponding position. We partition predictions into 10 evenly spaced confidence bins over $[0,100]$: $[0,10), [10,20), \ldots, [90,100]$.

For each bin $B_\ell$, we compute the average empirical accuracy and the average stated confidence of all predictions in that bin. ECE is then computed as the weighted average absolute gap between accuracy and confidence across bins:
\[
\mathrm{ECE}=\sum_{\ell=1}^{10}\frac{|B_\ell|}{N}
\left|\mathrm{acc}(B_\ell)-\mathrm{conf}(B_\ell)\right|,
\]
where $N$ is the total number of predictions, $|B_\ell|$ is the number of predictions in bin $B_\ell$, $\mathrm{acc}(B_\ell)$ is the percentage of correct predictions in the bin, and $\mathrm{conf}(B_\ell)$ is the average stated confidence in the bin. Lower ECE indicates better calibration.

In the main text, we report calibration primarily through calibration plots rather than ECE. This is because ECE is strongly negatively correlated with accuracy in our setting (Spearman's $\rho=-0.96$), making it less informative as a separate measure of calibration. Intuitively, models with lower accuracy tend to have larger confidence--accuracy gaps even when their confidence patterns are qualitatively similar. Calibration plots therefore provide a more interpretable view of whether models are overconfident or underconfident across confidence levels.

\section{Extracted Partial Description Validation\label{app:partial_description_validation}}

\paragraph{Metric definitions.}
Let $d$ denote the full description and let $A=\{a_1,\ldots,a_n\}$ denote the generated attribute list for an item. Text is normalized by lowercasing and tokenized with the regular expression \texttt{[A-Za-z0-9]+}, so punctuation is ignored and only alphanumeric tokens are retained.

For each attribute $a_i$, we compute an attribute support score $s(a_i,d)$ as the maximum of exact substring match and token overlap:
\[
s(a_i,d)=\max\left(\mathrm{substr}(a_i,d), \mathrm{overlap}(a_i,d)\right).
\]
An attribute is marked as supported if $s(a_i,d) \geq 0.45$.

The attribute support rate for an item is the fraction of generated attributes whose support score exceeds this threshold:
\[
\mathrm{support\_rate}(A,d)
=
\frac{
\sum_{i=1}^{n} \mathbb{1}[s(a_i,d) \geq 0.45]
}{
n
}.
\]

Description coverage measures how much of the original full description is recovered by the generated attributes:
\[
\mathrm{coverage}(A,d)
=
\frac{
\sum_{w \in V}
\min\left(c_A(w), c_d(w)\right)
}{
|T(d)|
}.
\]

For the selected underspecified expression, we compute the least-attribute support score as $s(a_{\mathrm{least}}, d)$, which verifies that the initial partial expression is grounded in the original description despite being intentionally underspecified.

\paragraph{Validation results.}
The extracted attributes are strongly supported by the original full descriptions. Across all items, the mean attribute support rate is 0.998, indicating that nearly all generated attributes can be matched back to the source description. The mean attribute support score is 0.980, and the mean description coverage is 0.895, suggesting that the decomposition preserves most of the descriptive content while keeping the attributes grounded in the original expression. The selected least-informative attributes are also well supported, with a mean least-attribute support score of 0.981.

These results suggest that our procedure produces source-grounded partial descriptions: the initial expressions used in \partialprotocol{} are intentionally underspecified, but not hallucinated or detached from the original human descriptions.

\section{Results\label{app:results}}




\subsection{LVLM and Human Matcher Performance Gap across Interaction Protocols \label{app:lvlm_human_gap}}

Tables~\ref{tab:appendix_gap_static}--\ref{tab:appendix_gap_none} report the full LVLM--human matcher accuracy gaps for each interaction protocol, complementing the best-over-protocol summary in Table~\ref{tab:lvlm_human_gap_heatmap}. Across \staticprotocol{}, \fullprotocol{}, and \partialprotocol{}, nearly all LVLMs significantly lag behind human matchers, although the size of the gap varies by model, dataset, and protocol. The only non-significant cases occur for Gemini-3.1-Pro, including one case in which it reaches parity with human matchers.

Under \oracleprotocol{}, all models significantly underperform human matchers. This comparison should be interpreted with caution, however, because the human reference data were not collected under an explicit \oracleprotocol{} setting. Instead, the human interactions more closely resemble a mixture of the other protocols, depending on how much detail human Directors provided in their object descriptions and how well Matchers understood those descriptions.

\begin{table}[]
\centering
\resizebox{\linewidth}{!}{%
\begin{tabular}{lrrrrr}
\toprule
Model & Dogs & Baskets 1 & Baskets 2 & Tangrams & All \\
\midrule
GPT-5.5 & \cellcolor{red!13}$-23.0^{***}$ & \cellcolor{red!10}$-16.0^{**}$ & \cellcolor{red!12}$-20.2^{***}$ & \cellcolor{red!27}$-57.7^{***}$ & \cellcolor{red!16}$-30.9^{***}$ \\
GPT-5.4 & \cellcolor{red!16}$-29.0^{***}$ & \cellcolor{red!13}$-22.0^{***}$ & \cellcolor{red!14}$-23.8^{***}$ & \cellcolor{red!22}$-44.6^{***}$ & \cellcolor{red!16}$-30.6^{***}$ \\
GPT-5.4-Mini & \cellcolor{red!20}$-41.0^{***}$ & \cellcolor{red!16}$-30.0^{***}$ & \cellcolor{red!22}$-45.8^{***}$ & \cellcolor{red!32}$-70.8^{***}$ & \cellcolor{red!24}$-48.8^{***}$ \\
Gemini-3.1-Pro & \cellcolor{red!9}$\mathbf{-12.0^{*}}$ & \cellcolor{red!8}$\mathbf{-9.0^{***}}$ & \cellcolor{red!8}$\mathbf{-10.7^{*}}$ & \cellcolor{red!6}$\mathbf{-6.0}$ & \cellcolor{red!8}$\mathbf{-9.2^{***}}$ \\
Gemini-3.1-Flash & \cellcolor{red!11}$-18.0^{***}$ & \cellcolor{red!13}$-23.0^{**}$ & \cellcolor{red!13}$-22.6^{***}$ & \cellcolor{red!23}$-47.0^{***}$ & \cellcolor{red!16}$-28.9^{***}$ \\
Gemma-4-31B-IT & \cellcolor{red!20}$-41.0^{***}$ & \cellcolor{red!20}$-40.0^{***}$ & \cellcolor{red!17}$-32.7^{***}$ & \cellcolor{red!24}$-48.8^{***}$ & \cellcolor{red!20}$-40.7^{***}$ \\
Qwen3-VL-32B & \cellcolor{red!16}$-31.0^{***}$ & \cellcolor{red!20}$-39.0^{***}$ & \cellcolor{red!21}$-41.7^{***}$ & \cellcolor{red!33}$-72.6^{***}$ & \cellcolor{red!23}$-47.9^{***}$ \\
Qwen3-VL-8B & \cellcolor{red!23}$-47.0^{***}$ & \cellcolor{red!26}$-56.0^{***}$ & \cellcolor{red!32}$-69.0^{***}$ & \cellcolor{red!34}$-73.8^{***}$ & \cellcolor{red!29}$-63.1^{***}$ \\
\bottomrule
\end{tabular}
}
\caption{Mean LVLM--human matcher accuracy gap (\%) under the \staticprotocol{} protocol. Darker red denotes larger deficits relative to human matchers, and bold marks the smallest gap within each dataset. Significance is assessed with paired $t$-tests: $^{*}p \le .05$, $^{**}p \le .01$, $^{***}p \le .001$.}
\label{tab:appendix_gap_static}
\end{table}

\begin{table}[]
\centering
\resizebox{\linewidth}{!}{%
\begin{tabular}{lrrrrr}
\toprule
Model & Dogs & Baskets 1 & Baskets 2 & Tangrams & All \\
\midrule
GPT-5.5 & \cellcolor{red!11}$-18.0^{***}$ & \cellcolor{red!10}$-15.0^{**}$ & \cellcolor{red!11}$-18.5^{***}$ & \cellcolor{red!25}$-51.2^{***}$ & \cellcolor{red!15}$-27.2^{***}$ \\
GPT-5.4 & \cellcolor{red!14}$-26.0^{***}$ & \cellcolor{red!10}$-16.0^{***}$ & \cellcolor{red!12}$-20.8^{***}$ & \cellcolor{red!21}$-42.9^{***}$ & \cellcolor{red!15}$-27.3^{***}$ \\
GPT-5.4-Mini & \cellcolor{red!20}$-40.0^{***}$ & \cellcolor{red!18}$-35.0^{***}$ & \cellcolor{red!17}$-32.1^{***}$ & \cellcolor{red!35}$-78.0^{***}$ & \cellcolor{red!23}$-47.7^{***}$ \\
Gemini-3.1-Pro & \cellcolor{red!7}$\mathbf{-7.0^{*}}$ & \cellcolor{red!5}$\mathbf{-3.0}$ & \cellcolor{red!7}$\mathbf{-8.3^{*}}$ & \cellcolor{red!7}$\mathbf{-7.7^{*}}$ & \cellcolor{red!7}$\mathbf{-6.8^{***}}$ \\
Gemini-3.1-Flash & \cellcolor{red!12}$-19.0^{***}$ & \cellcolor{red!16}$-30.0^{***}$ & \cellcolor{red!14}$-24.4^{***}$ & \cellcolor{red!23}$-48.2^{***}$ & \cellcolor{red!17}$-31.4^{***}$ \\
Gemma-4-31B-IT & \cellcolor{red!19}$-37.0^{***}$ & \cellcolor{red!16}$-31.0^{***}$ & \cellcolor{red!15}$-27.4^{***}$ & \cellcolor{red!24}$-50.0^{***}$ & \cellcolor{red!19}$-36.7^{***}$ \\
Qwen3-VL-32B & \cellcolor{red!14}$-25.0^{***}$ & \cellcolor{red!14}$-26.0^{**}$ & \cellcolor{red!21}$-42.3^{***}$ & \cellcolor{red!36}$-79.8^{***}$ & \cellcolor{red!23}$-46.2^{***}$ \\
Qwen3-VL-8B & \cellcolor{red!21}$-42.0^{***}$ & \cellcolor{red!26}$-55.0^{***}$ & \cellcolor{red!30}$-66.1^{***}$ & \cellcolor{red!35}$-76.2^{***}$ & \cellcolor{red!29}$-61.7^{***}$ \\
\bottomrule
\end{tabular}
}
\caption{Mean LVLM--human matcher accuracy gap (\%) under the \fullprotocol{} protocol. Darker red denotes larger deficits relative to human matchers, and bold marks the smallest gap within each dataset. Significance is assessed with paired $t$-tests: $^{*}p \le .05$, $^{**}p \le .01$, $^{***}p \le .001$.}
\label{tab:appendix_gap_full}
\end{table}

\begin{table}[]
\centering
\resizebox{\linewidth}{!}{%
\begin{tabular}{lrrrrr}
\toprule
Model & Dogs & Baskets 1 & Baskets 2 & Tangrams & All \\
\midrule
GPT-5.5 & \cellcolor{red!10}$-14.0^{**}$ & \cellcolor{red!10}$-15.0^{**}$ & \cellcolor{red!14}$-25.0^{***}$ & \cellcolor{red!20}$-41.1^{***}$ & \cellcolor{red!14}$-25.3^{***}$ \\
GPT-5.4 & \cellcolor{red!14}$-26.0^{***}$ & \cellcolor{red!13}$-23.0^{***}$ & \cellcolor{red!13}$-22.0^{***}$ & \cellcolor{red!24}$-50.0^{***}$ & \cellcolor{red!16}$-31.2^{***}$ \\
GPT-5.4-Mini & \cellcolor{red!29}$-62.0^{***}$ & \cellcolor{red!20}$-40.0^{***}$ & \cellcolor{red!26}$-56.0^{***}$ & \cellcolor{red!34}$-75.6^{***}$ & \cellcolor{red!28}$-59.6^{***}$ \\
Gemini-3.1-Pro & \cellcolor{red!7}$\mathbf{-7.0}$ & $\mathbf{0.0}$ & \cellcolor{red!7}$\mathbf{-8.3^{*}}$ & \cellcolor{red!9}$\mathbf{-13.1^{**}}$ & \cellcolor{red!7}$\mathbf{-7.7^{***}}$ \\
Gemini-3.1-Flash & \cellcolor{red!17}$-32.0^{***}$ & \cellcolor{red!16}$-29.0^{***}$ & \cellcolor{red!20}$-40.5^{***}$ & \cellcolor{red!28}$-60.7^{***}$ & \cellcolor{red!21}$-42.2^{***}$ \\
Gemma-4-31B-IT & \cellcolor{red!20}$-41.0^{***}$ & \cellcolor{red!17}$-32.0^{***}$ & \cellcolor{red!18}$-34.5^{***}$ & \cellcolor{red!22}$-44.6^{***}$ & \cellcolor{red!19}$-38.3^{***}$ \\
Qwen3-VL-32B & \cellcolor{red!17}$-32.0^{***}$ & \cellcolor{red!18}$-36.0^{***}$ & \cellcolor{red!24}$-50.6^{***}$ & \cellcolor{red!37}$-81.5^{***}$ & \cellcolor{red!25}$-52.7^{***}$ \\
Qwen3-VL-8B & \cellcolor{red!25}$-53.0^{***}$ & \cellcolor{red!26}$-54.0^{***}$ & \cellcolor{red!29}$-61.9^{***}$ & \cellcolor{red!33}$-72.0^{***}$ & \cellcolor{red!29}$-61.4^{***}$ \\
\bottomrule
\end{tabular}
}
\caption{Mean LVLM--human matcher accuracy gap (\%) under the \partialprotocol{} protocol. Darker red denotes larger deficits relative to human matchers, and bold marks the smallest gap within each dataset. Significance is assessed with paired $t$-tests: $^{*}p \le .05$, $^{**}p \le .01$, $^{***}p \le .001$.}
\label{tab:appendix_gap_partial}
\end{table}

\begin{table}[t]
\centering

\resizebox{\linewidth}{!}{%
\begin{tabular}{lrrrrr}
\toprule
Model & Dogs & Baskets 1 & Baskets 2 & Tangrams & All \\
\midrule
GPT-5.5 & \cellcolor{red!14}$\mathbf{-26.0^{***}}$ & \cellcolor{red!14}$-25.0^{***}$ & \cellcolor{red!24}$-50.0^{***}$ & \cellcolor{red!20}$\mathbf{-40.5^{***}}$ & \cellcolor{red!19}$\mathbf{-37.0^{***}}$ \\
GPT-5.4 & \cellcolor{red!23}$-48.0^{***}$ & \cellcolor{red!27}$-58.0^{***}$ & \cellcolor{red!25}$-51.8^{***}$ & \cellcolor{red!26}$-56.0^{***}$ & \cellcolor{red!25}$-53.5^{***}$ \\
GPT-5.4-Mini & \cellcolor{red!32}$-70.0^{***}$ & \cellcolor{red!36}$-81.0^{***}$ & \cellcolor{red!39}$-87.5^{***}$ & \cellcolor{red!40}$-89.9^{***}$ & \cellcolor{red!37}$-83.2^{***}$ \\
Gemini-3.1-Pro & \cellcolor{red!17}$-33.0^{***}$ & \cellcolor{red!12}$\mathbf{-19.0^{**}}$ & \cellcolor{red!22}$\mathbf{-44.0^{***}}$ & \cellcolor{red!26}$-56.0^{***}$ & \cellcolor{red!20}$-40.0^{***}$ \\
Gemini-3.1-Flash & \cellcolor{red!26}$-55.0^{***}$ & \cellcolor{red!25}$-53.0^{***}$ & \cellcolor{red!27}$-57.1^{***}$ & \cellcolor{red!31}$-67.3^{***}$ & \cellcolor{red!28}$-58.8^{***}$ \\
Gemma-4-31B-IT & \cellcolor{red!30}$-65.0^{***}$ & \cellcolor{red!30}$-66.0^{***}$ & \cellcolor{red!30}$-64.9^{***}$ & \cellcolor{red!37}$-82.7^{***}$ & \cellcolor{red!32}$-70.3^{***}$ \\
Qwen3-VL-32B & \cellcolor{red!29}$-63.0^{***}$ & \cellcolor{red!30}$-65.0^{***}$ & \cellcolor{red!35}$-78.6^{***}$ & \cellcolor{red!36}$-79.8^{***}$ & \cellcolor{red!33}$-72.8^{***}$ \\
Qwen3-VL-8B & \cellcolor{red!28}$-61.0^{***}$ & \cellcolor{red!34}$-76.0^{***}$ & \cellcolor{red!38}$-83.9^{***}$ & \cellcolor{red!34}$-75.6^{***}$ & \cellcolor{red!34}$-75.1^{***}$ \\
\bottomrule
\end{tabular}
}
\caption{Mean LVLM--human matcher accuracy gap (\%) under the \oracleprotocol{} protocol. Darker red denotes larger deficits relative to human matchers, and bold marks the smallest gap within each dataset. Significance is assessed with paired $t$-tests: $^{*}p \le .05$, $^{**}p \le .01$, $^{***}p \le .001$.}
\label{tab:appendix_gap_none}
\end{table}

\begin{table}[]
\centering
\resizebox{1\linewidth}{!}{%
\begin{tabular}{llll}
\toprule
Model & \fullprotocol{} & \partialprotocol{} & Total \\
\midrule
GPT-5.5 & 7 & 10 & 17 \\
GPT-5.4 & 4 & 2 & 6 \\
GPT-5.4-Mini & 0 & 1 & 1 \\
Gemini-3.1-Pro & 9 & 8 & 17 \\
Gemini-3.1-Flash & 1 & 4 & 5 \\
Gemma-4-31B-IT & 2 & 1 & 3 \\
\bottomrule
\end{tabular}
}
\caption{Distribution of self-repair cases by model and interaction protocol. Counts indicate the number of observed self-repairs under \fullprotocol{} and \partialprotocol{}. We did not observe self-repairs under \oracleprotocol{}.}
\label{tab:self_repair_count_distribution}
\end{table}

\begin{table*}[]
\centering
\resizebox{\linewidth}{!}{%
\begin{tabular}{llllll}
\toprule
Model & Count & Wrong $\rightarrow$ Correct & Correct $\rightarrow$ Wrong & Wrong $\rightarrow$ Wrong & $\Delta$ Confidence \\
\midrule
GPT-5.5 & 17 & 47.1 & 5.9 & 47.1 & $-14.9$ \\
GPT-5.4 & 6 & 0.0 & 50.0 & 50.0 & $+11.2$ \\
GPT-5.4-Mini & 1 & 100.0 & 0.0 & 0.0 & $-15.0$ \\
Gemini-3.1-Pro & 17 & 76.5 & 0.0 & 23.5 & $-4.3$ \\
Gemini-3.1-Flash & 5 & 60.0 & 20.0 & 20.0 & $+1.0$ \\
Gemma-4-31B-IT & 3 & 33.3 & 0.0 & 66.7 & $-16.7$ \\
\bottomrule
\end{tabular}
}
\caption{Self-repair outcomes by model. Count denotes the number of self-repair cases. Transition columns report the percentage of repairs that change an initially wrong selection to correct, correct selection to wrong, or wrong selection to another wrong answer. $\Delta$ Confidence denotes the mean change in confidence after repair; positive values indicate increased confidence, and negative values indicate decreased confidence.}
\label{tab:self_repair_outcomes}
\end{table*}

\subsection{Self-Repair Analysis\label{app:self-repair}}

Tables~\ref{tab:self_repair_outcomes} and~\ref{tab:self_repair_count_distribution} summarize self-repair behavior for models that revise their selections at least once. Self-repairs are relatively rare, accounting for no more than 2\% of the 536 matches in any interaction protocol, and occur only under \fullprotocol{} and \partialprotocol{}, with no observed cases under \oracleprotocol{}. When models do revise, outcomes vary substantially: some repairs change initially wrong selections to correct ones, especially for Gemini-3.1-Pro, while others introduce errors or replace one wrong answer with another, as in about half of GPT-5.5 cases. Confidence changes are also mixed, suggesting that revisions do not always correspond to increased certainty. Overall, self-repair is observable but limited in current LVLM matchers.

\begin{figure*}[!htb]
    \centering    \includegraphics[width=1\linewidth]{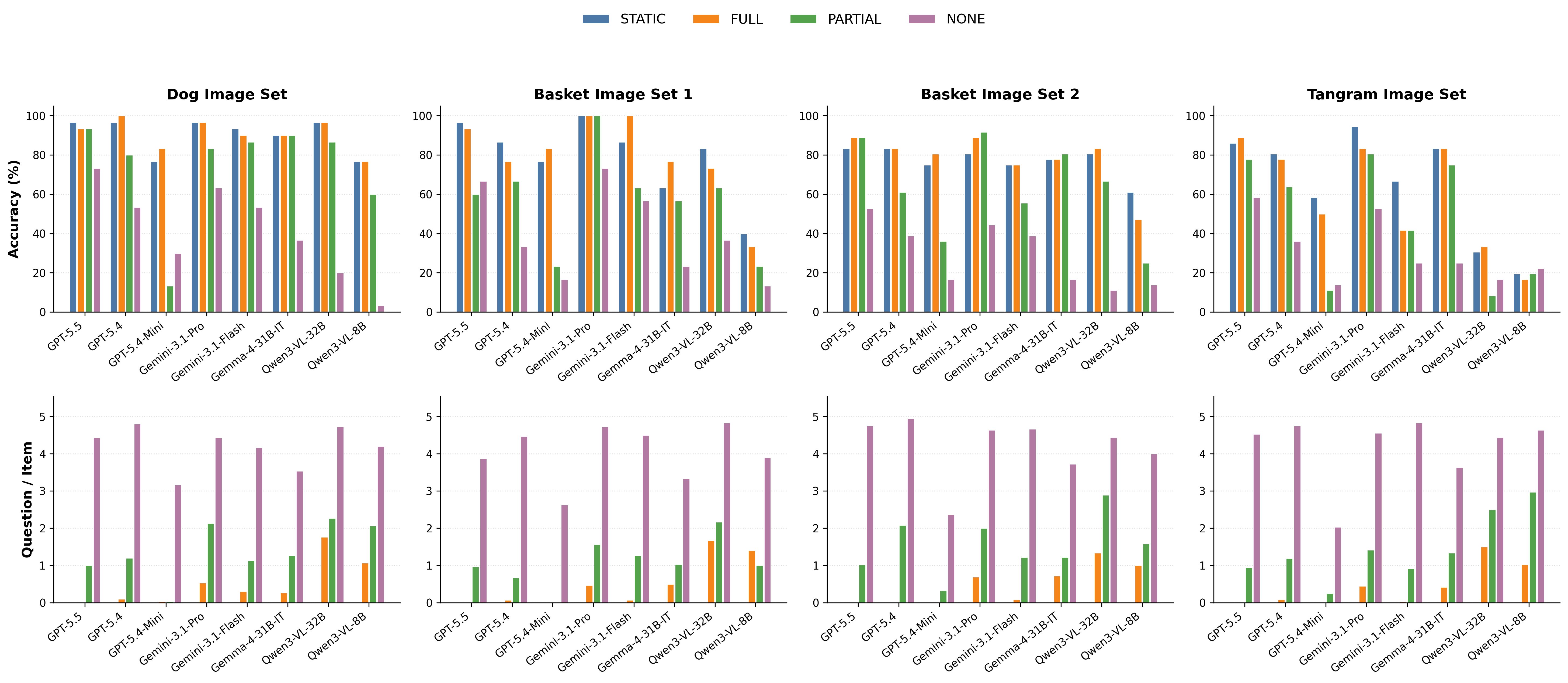}
    \caption{Overall results across the four image sets from the follow-up experiments with AI-generated descriptions. Top: average matcher accuracy. 
    Bottom: average questions per target item. Colors denote the four interaction protocols; horizontal lines mark human and random baselines.}
    \label{fig:overall_results_ai_generated_descriptions}
\end{figure*}

\begin{figure}[]
    \centering    \includegraphics[width=1\linewidth]{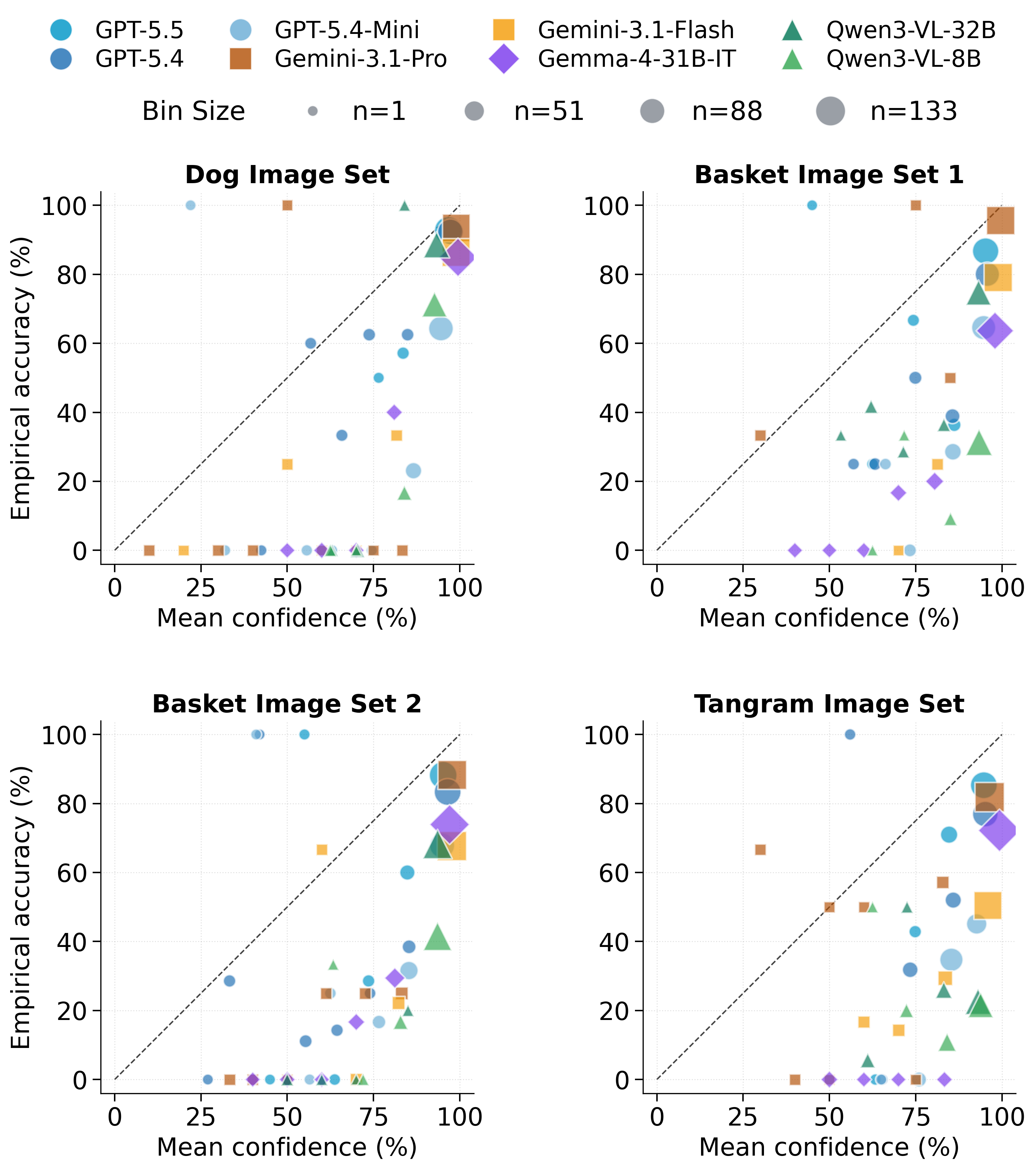}
    \caption{Protocol-level calibration from the follow-up experiments with AI-generated descriptions. Points show model-specific confidence bins from 10 evenly spaced bins between 0 and 100; point size is proportional to bin count. The dashed diagonal denotes perfect calibration.}
    \label{fig:dataset_level_confidence_calibration_ai_generated_descriptions}
\end{figure}

\begin{figure}[]
    \centering    \includegraphics[width=\linewidth]{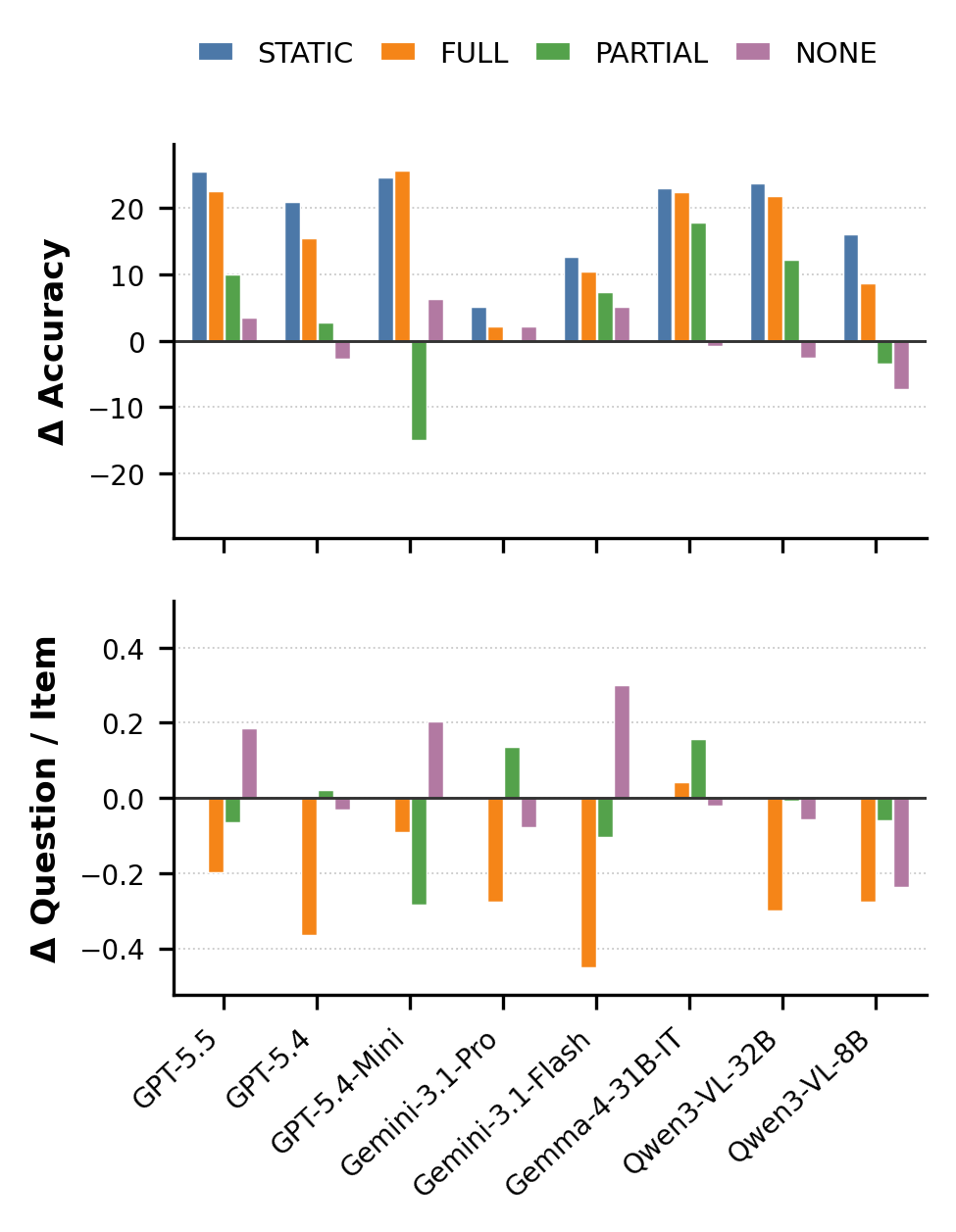}
    \caption{Differences in LVLM outcomes and question-asking behavior between AI- and human-generated descriptions, averaged across all four datasets. Bars show $\Delta=$ AI-generated $-$ human-generated descriptions for accuracy and questions per item.}\label{fig:overall_human_versus_ai_all_datasets}
\end{figure}

\section{Follow-Up Studies\label{app:follow-up-studies}} 

This section provides additional implementation details and results for the follow-up studies conducted in Section~\ref{sec:follw-up-studies}.

\subsection{AI-Generated Descriptions}

\paragraph{Experimental Procedure} For experiments on AI-generated descriptions, we replace human-generated descriptions with GPT-5.4-generated ones and rerun the main experiments across all models and interaction protocols. We repeat each experiment three times using the same description map (see below) but with shuffled description order. This controls for variation in item ordering while isolating the effect of description source, allowing us to test whether the main results depend on the use of human referring expressions.

\paragraph{Additional Result Discussion}
Figure~\ref{fig:overall_results_ai_generated_descriptions} reports the full results for each dataset. The overall patterns are comparable to those in the main experiments with human-generated descriptions (Figure~\ref{fig:overallResults}; Section~\ref{sec:results}). LVLMs also remain overconfident and undercalibrated even with AI-generated descriptions, as shown in Figure~\ref{fig:dataset_level_confidence_calibration_ai_generated_descriptions}.

For easier comparison, we aggregate results across the four datasets and visualize the differences in accuracy and question-asking behavior between AI- and human-generated descriptions in Figure~\ref{fig:overall_human_versus_ai_all_datasets}. Figure~\ref{fig:overall_human_versus_ai_all_datasets} shows that LVLMs generally achieve higher accuracy with AI-generated descriptions than with human-generated descriptions, especially under \staticprotocol{} and \fullprotocol{}. Models also tend to ask fewer questions under \fullprotocol{}, suggesting that AI-generated descriptions are easier for LVLMs to interpret. This contrast indicates that human referring expressions pose a distinct challenge, even when they are sufficiently informative for human matchers. Importantly, however, the overall protocol-level patterns remain similar: \staticprotocol{} and \fullprotocol{} are strongest, \partialprotocol{} is more variable, and \oracleprotocol{} remains difficult.

\begin{table}[t]
\centering
\begin{tabular}{lll}
\toprule
Reasoning & GPT-5.4 & GPT-5.4-Mini \\
\midrule
None   
& $0.6$ 
& $0.2$ \\

Medium 
& $0.8_{\scriptscriptstyle \textcolor{ForestGreen}{+0.2}}$ 
& $0.1_{\scriptscriptstyle \textcolor{red}{-0.1}}$ \\

High   
& $0.9_{\scriptscriptstyle \textcolor{ForestGreen}{+0.3}}$ 
& $0.0_{\scriptscriptstyle \textcolor{red}{-0.2}}$ \\
\bottomrule
\end{tabular}
\caption{Average number of questions per target item under \fullprotocol{} for different reasoning-effort settings. Subscripts report differences relative to the default no-reasoning setting; \textcolor{ForestGreen}{green} indicates increases and \textcolor{red}{red} indicates decreases.}
\label{tab:reasoning_effort_questions}
\end{table}

\begin{figure}[]
    \centering    \includegraphics[width=1\linewidth]{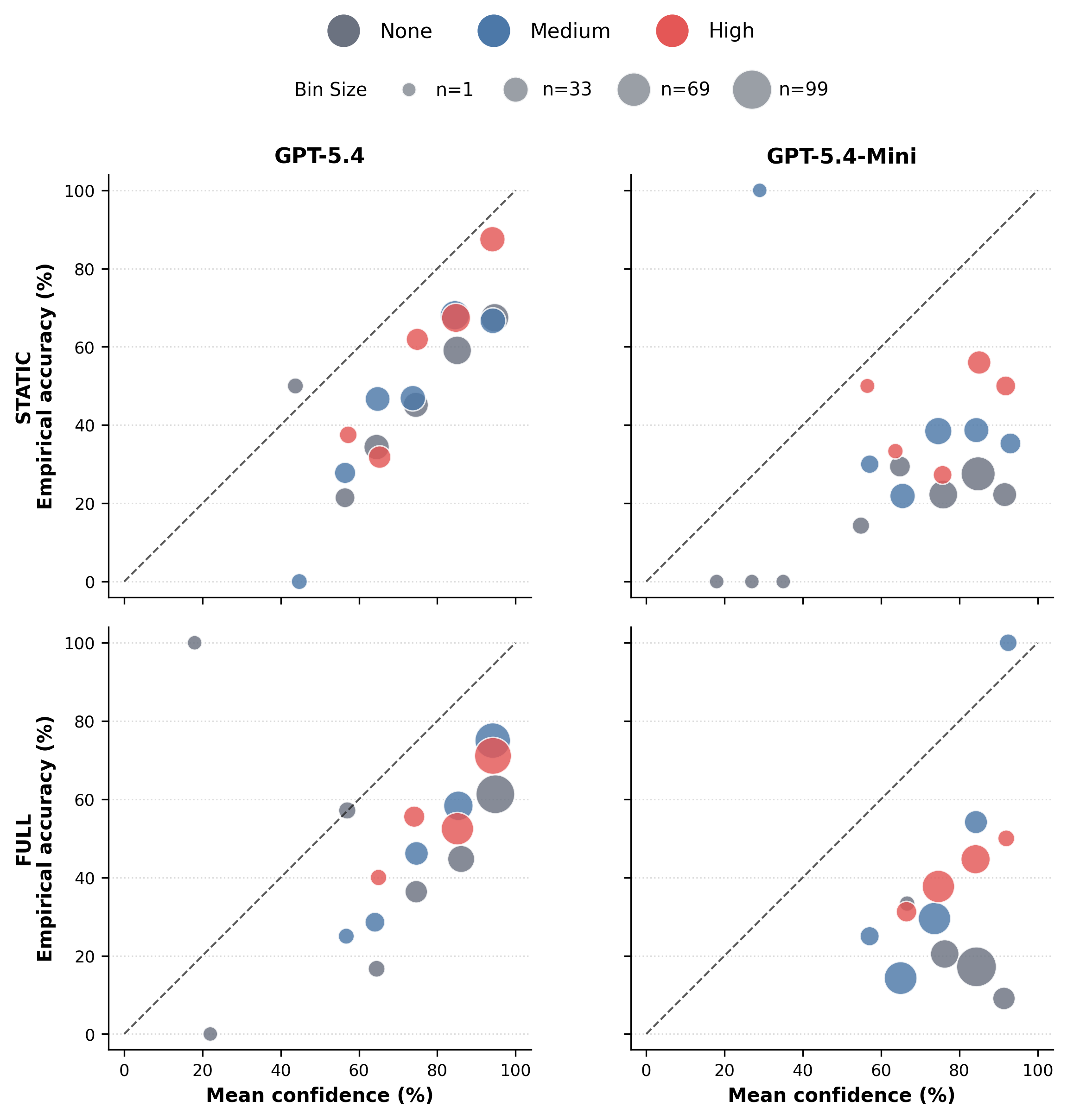}
    \caption{Calibration under different reasoning-effort settings for GPT-5.4 and GPT-5.4-Mini. Each point represents a confidence bin, with the $x$-axis showing mean reported confidence and the $y$-axis showing empirical accuracy. Point size indicates bin size, colors denote reasoning effort, and the dashed diagonal marks perfect calibration.}
    \label{fig:reasoning_effort_confidence_calibration}
\end{figure}

\subsection{Reasoning Effort\label{app:reasoning_effort}}

\paragraph{Reasoning-effort setup.}
Default reasoning effort is model-specific rather than uniform across GPT models. GPT-5.5 uses medium reasoning effort by default, whereas GPT-5.4 defaults to no reasoning. We therefore vary the reasoning effort of GPT-5.4 and GPT-5.4-Mini through the official Responses API to isolate the effect of test-time reasoning.

\paragraph{Reasoning effort changes question asking modestly.}
Table~\ref{tab:reasoning_effort_questions} shows that higher reasoning effort slightly increases question asking for GPT-5.4 under \fullprotocol{}, while GPT-5.4-Mini asks very few questions across settings. Together with Table~\ref{tab:reasoning_effort_accuracy}, this suggests that reasoning improves interactive performance partly by supporting information seeking, but not simply by increasing question volume.

\paragraph{Reasoning effort slightly improves calibration, but gains plateau from medium to high.}
Figure~\ref{fig:reasoning_effort_confidence_calibration} shows that higher reasoning effort generally brings more calibration bins closer to the diagonal, especially when accounting for bin sizes, indicating slightly better alignment between confidence and accuracy. However, the shift from medium to high reasoning is small, suggesting that scaling reasoning effort does not reliably improve calibration. Most bins still fall below the diagonal, so both models remain overconfident.

\begin{figure}[]
    \centering    \includegraphics[width=1.0\linewidth]{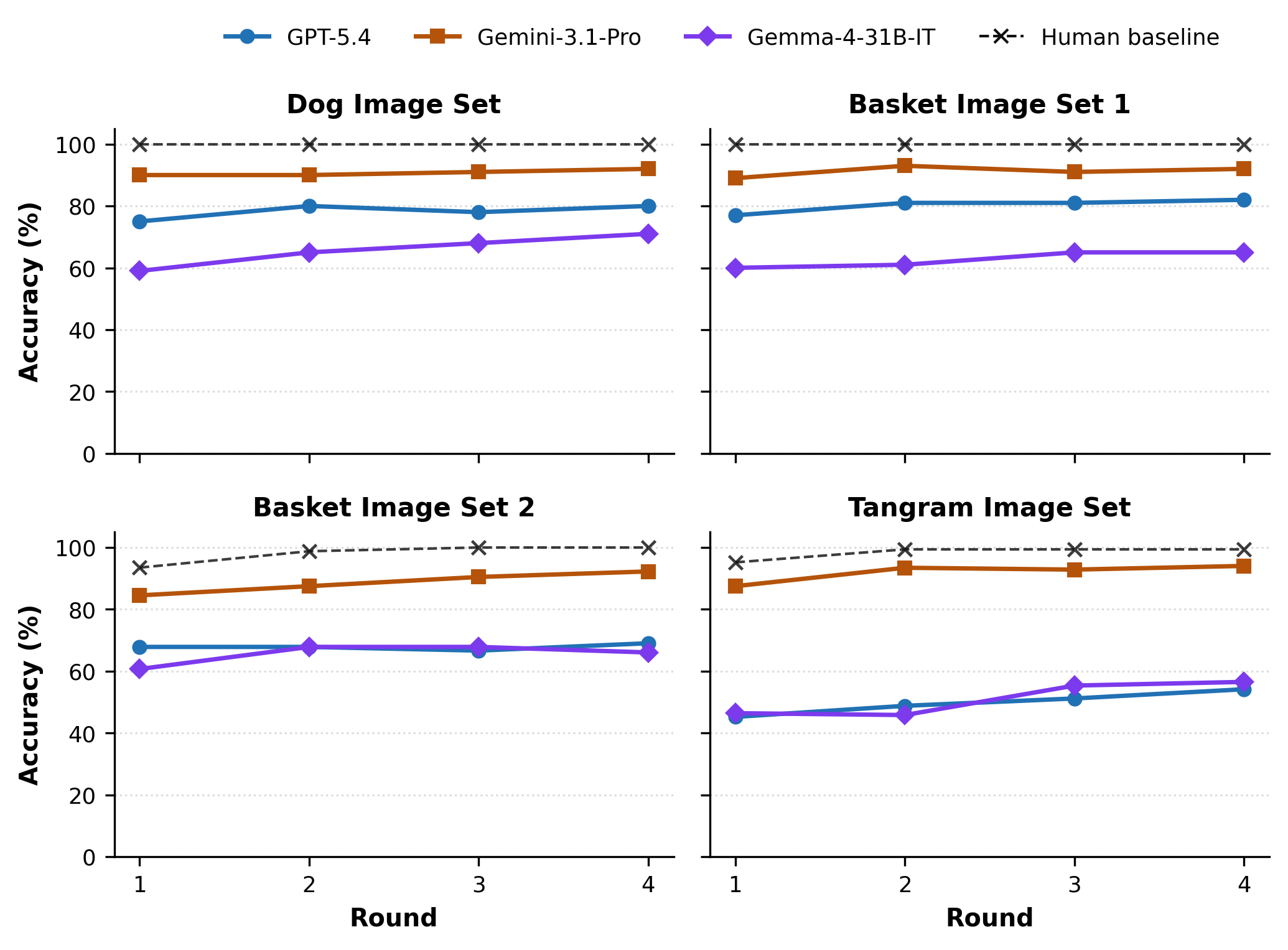}
    \caption{Accuracy (\%) across rounds on the four image sets under \staticprotocol{}.}\label{fig:static_multi_round_accuracy_by_dataset_2x2}
\end{figure}

\begin{figure}[]
    \centering    \includegraphics[width=1.0\linewidth]{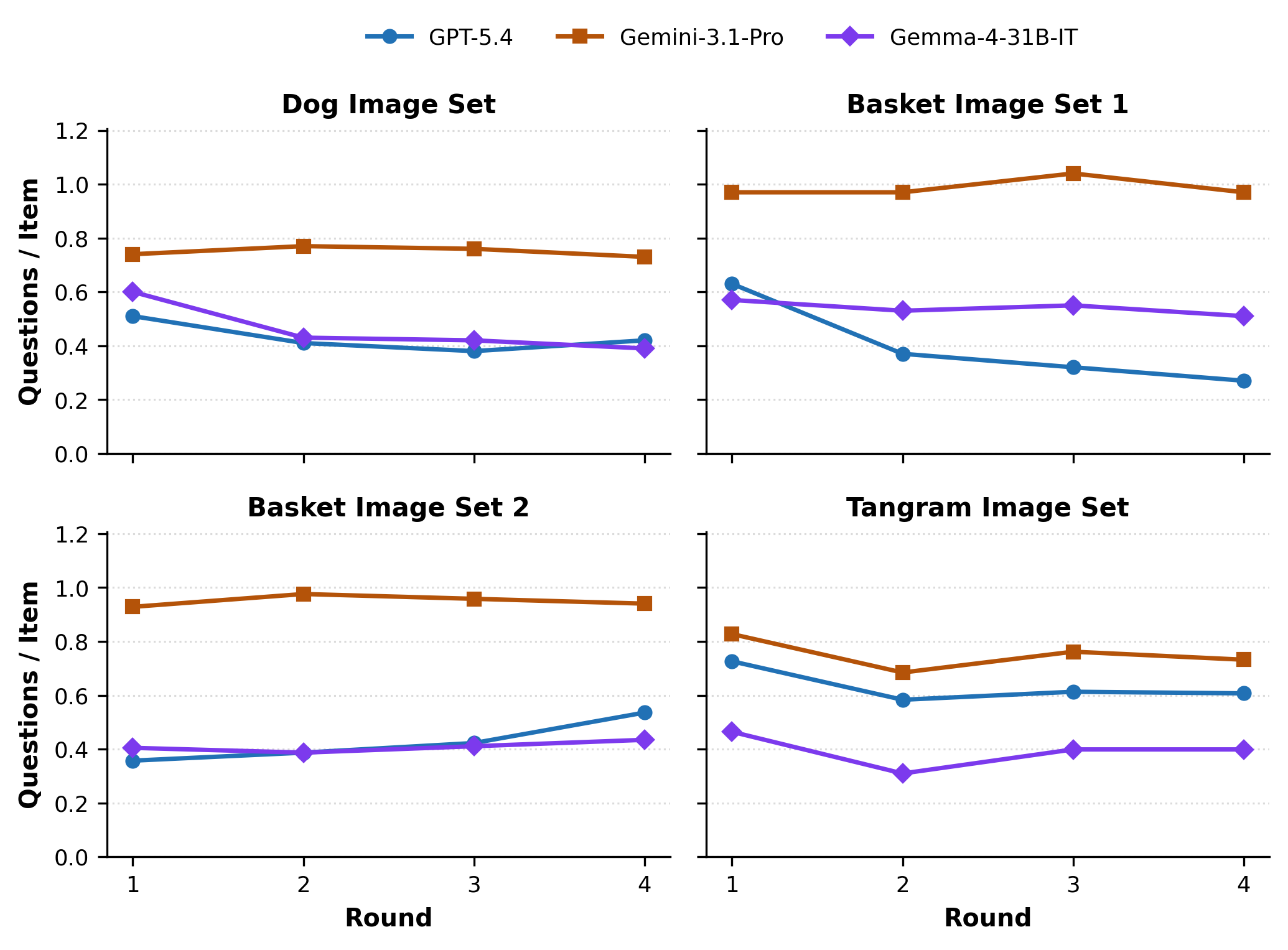}
    \caption{Average number of questions per item across rounds on the tangrams under \fullprotocol{}.}\label{fig:full_multi_round_questions_by_dataset_2x2}
\end{figure}

\begin{figure*}[]
    \centering    \includegraphics[width=1\linewidth]{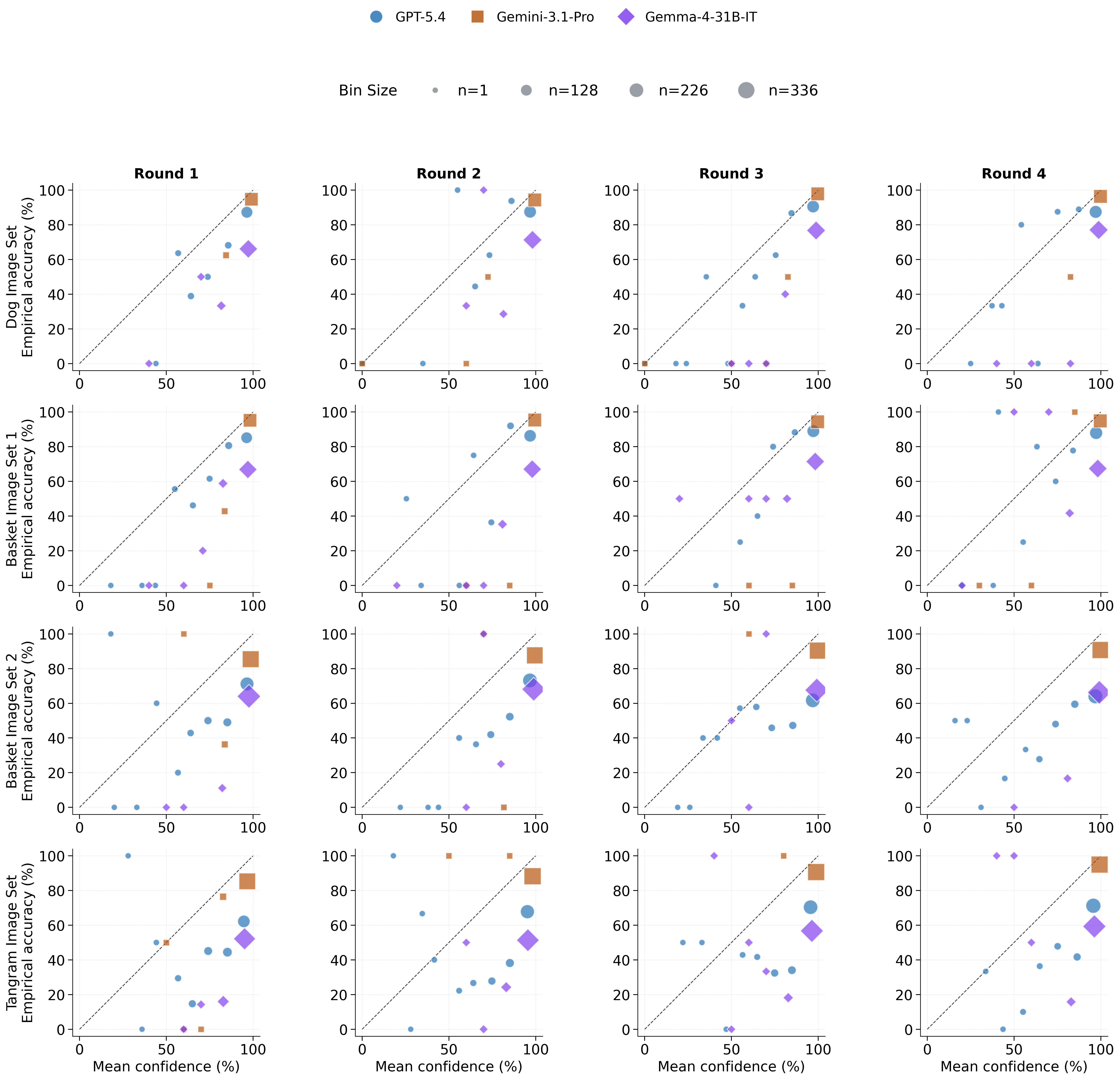}
    \caption{Round-level calibration across datasets in the four-round follow-up experiments. Points show model-specific confidence bins from 10 evenly spaced bins between 0 and 100; point size is proportional to bin count. The dashed diagonal denotes perfect calibration.}
    \label{fig:multi_round_dataset_round_calibration_4x4}
\end{figure*}

\subsection{Multi-Round Repeated References\label{app:multi-round}}

\paragraph{Per-dataset accuracy results under \staticprotocol{}.}
Figure~\ref{fig:static_multi_round_accuracy_by_dataset_2x2} shows model performance on tangrams across rounds under \staticprotocol{}. Models improve matching performance even without the ability to ask questions, confirming that they can comprehend the increasingly efficient referring expressions generated by humans over repeated interaction.

\paragraph{Question asking under \fullprotocol{}.}
Figure~\ref{fig:full_multi_round_questions_by_dataset_2x2} shows the average number of questions per item across rounds under \fullprotocol{}. Question rates are generally stable or decrease slightly over rounds for most models and datasets. 

\paragraph{Model calibration across rounds.}
Figure~\ref{fig:multi_round_dataset_round_calibration_4x4} shows round-level calibration across datasets, aggregating results from \staticprotocol{} and \fullprotocol{} because the two protocols exhibit broadly similar patterns. Models remain imperfectly calibrated across rounds: many confidence bins fall below the diagonal, indicating overconfidence. Calibration does not consistently improve with repeated exposure, even when accuracy increases over rounds. This suggests that LVLMs can sometimes benefit from repeated references without reliably improving their uncertainty estimates.

\subsection{Alternative Director Robustness\label{app:alternative-director}}

\paragraph{Experimental procedure.}
To test whether Matcher results depend on one simulated information source, we replace GPT-5.4 with Gemini-3.1-Flash as the Director and rerun \fullprotocol{} and \partialprotocol{} on Basket Image Set 2 and Tangrams. We evaluate both human- and GPT-5.4-generated initial descriptions and otherwise preserve the original prompts, candidate views, and turn budget.

\begin{table*}[t]
\centering
\small
\setlength{\tabcolsep}{5pt}
\begin{tabular}{lrrrr}
\toprule
& \multicolumn{2}{c}{GPT-5.4 Director} & \multicolumn{2}{c}{Gemini-3.1-Flash Director} \\
\cmidrule(lr){2-3}\cmidrule(lr){4-5}
Matcher & \fullprotocol{} & \partialprotocol{} & \fullprotocol{} & \partialprotocol{} \\
\midrule
GPT-5.5            & 59.5 & 61.3 & 59.5 & 55.7 \\
GPT-5.4            & 62.8 & 58.3 & 57.4 & 51.2 \\
GPT-5.4-Mini       & 39.3 & 28.6 & 38.7 & 28.0 \\
Gemini-3.1-Pro     & 81.5 & 83.6 & 86.3 & 85.1 \\
Gemini-3.1-Flash   & 58.0 & 43.8 & 58.3 & 37.5 \\
Gemma-4-31B-IT     & 54.5 & 54.8 & 53.9 & 46.4 \\
Qwen3-VL-32B       & 33.3 & 28.3 & 33.0 & 33.0 \\
Qwen3-VL-8B        & 23.2 & 27.4 & 16.1 & 21.1 \\
\bottomrule
\end{tabular}
\caption{Accuracy (\%) with GPT-5.4 and Gemini-3.1-Flash Directors, aggregated across Basket Image Set 2 and Tangrams and across human- and AI-generated descriptions. Matcher rankings across Directors remain highly correlated ($\rho=.929$ for \fullprotocol{} and $.976$ for \partialprotocol{}).}
\label{tab:alternative_director}
\end{table*}

\paragraph{Results.}
Table~\ref{tab:alternative_director} shows that individual model--protocol combinations vary with the Director, as expected in a collaborative task, but Matcher rankings remain highly stable. Rank correlations between results with GPT-5.4 and Gemini-3.1-Flash Directors are $\rho=.929$ under \fullprotocol{} and $\rho=.976$ under \partialprotocol{}. GPT-5.4 is generally the stronger Director, yet the main protocol-level patterns and Matcher hierarchy do not depend on using GPT-5.4 alone. The correlations are computed over Matcher-level accuracies aggregated across the two datasets and description sources.

\subsection{Face-Domain Robustness}\label{app:face-domain}

\paragraph{Experimental procedure.}
We evaluate five representative Matchers on six image sets from the public FACES database \citep{Holland2018}. Each set contains 12 images of one individual displaying varied facial expressions. Because these image sets do not include human interaction data, we generate initial descriptions with GPT-5.4 and evaluate the same four protocols. Director and Matcher indices are randomized independently as in the main study. This experiment contains no protocol-matched human baseline and is intended as a robustness check in another fine-grained within-category context.

\begin{table}[t]
\centering
\small

\resizebox{\linewidth}{!}{%
\begin{tabular}{lrrrr}
\toprule
Matcher & \staticprotocol{} & \fullprotocol{} & \partialprotocol{} & \oracleprotocol{} \\
\midrule
GPT-5.4          & 53.7 & 55.1 & 56.9 & 52.3 \\
GPT-5.4-Mini     & 44.9 & 47.7 & 26.9 & 16.2 \\
Gemini-3.1-Flash & 55.6 & 50.0 & 38.0 & 17.6 \\
Gemma-4-31B-IT   & 55.1 & 54.6 & 54.2 & 49.4 \\
Qwen3-VL-8B      & 34.3 & 31.5 & 33.3 & 20.8 \\
\midrule
Overall          & 48.7 & 47.8 & 41.9 & 30.7 \\
\bottomrule
\end{tabular}
}
\caption{Accuracy (\%) on six face image sets with AI-generated descriptions. Each image set contains 12 images. There is no protocol-matched human baseline for this robustness experiment.}
\label{tab:face_domain}
\end{table}

\paragraph{Results and scope.}
Table~\ref{tab:face_domain} reproduces the main aggregate pattern: \staticprotocol{} and \fullprotocol{} are comparable (48.7\% and 47.8\%), \partialprotocol{} is lower (41.9\%), and \oracleprotocol{} is most difficult (30.7\%). The result extends the protocol-level observation beyond dogs, baskets, and tangrams, but faces remain a controlled within-category matching context; this study does not establish generality to cluttered scenes, spatial relations, or open-world visual search.

\onecolumn
\section{Prompts for Visual Grounding Experiments}\label{app:prompts}

This section reports the prompts used in the study for the visual grounding experiments. 

\subsection{Task Overview}

We reuse the same task overview prompt for both the Director and Matcher across all interaction protocols to maximize the comparability across the four interaction protocols.

\begin{promptbox}
### Task Overview

You will participate in a visual reference task involving two roles: Director and Matcher. The two participants may not share the same visual context.

Both participants are shown randomized sets of indexed visual items. These visual items may be objects, scenes, or other image-based stimuli. The participants share a subset of target items, but the Matcher may also see additional distractor items that are not visible to the Director.

The goal is for the Director and Matcher to collaborate so that the Matcher can identify each target item as accurately as possible.

The task may consist of multiple rounds. Please strictly follow the instructions for your assigned role across all rounds.
\end{promptbox}

\subsection{Matcher Output Format Instructions}

Across all interaction protocols, Matcher is prompted to produce a JSON object containing the target item index and its confidence level (\%) rated from 0 to 100. 

The output format instruction for the \staticprotocol{} protocol is as follows.

\begin{promptbox}
### Output Format Instruction

Output your response in JSON format using the following structure:

{
    "target_item_index": 5,
    "confidence": 75
}

**Field Definitions**

- "target_item_index": the index of the item you are selecting as the match for the identified target item.
- "confidence": your confidence level (%) in the selection, rated from 0 (not confident at all) to 100 (most confident).
\end{promptbox}

The output format instruction for the other three interactive protocols (\fullprotocol{}, \partialprotocol{}, and \oracleprotocol{}) is as follows.

\begin{promptbox}
### Output Format Instruction

Output your response in JSON format using the following structure:

{
  "utterance": "Your natural, concise utterance to the Director.",
  "selections": [
    {
      "nth_item": 2,
      "target_item_index": 5,
      "confidence": 75
    },
    ...
  ]
}

**Field Definitions**

- "utterance" should contain your natural language utterance you say to the Director. Do not mention indices or positional cues here.
- "selections" should be a list of items you are selecting based on the Director's responses. Each item selection should include:
  - "nth_item": the order of the target item being identified (starting from 1 for the first item identified in the current round).
  - "target_item_index": the index of the item you are selecting as the match for the identified target item.
  - "confidence": your confidence level (%) in the selection, rated from 0 (not confident at all) to 100 (most confident).

**Output Rules**

- If you do not have enough information to make a selection, set "selections" to an empty list.
- If you have enough information to make a selection for the current item, include that item in "selections".
- If you are revising one or more earlier selections, include each revised item as a separate entry in "selections".
- If you are both revising earlier selections and selecting the current item, include all of those updates in "selections" in the same response.
- Unless you are revising an earlier choice, "selections" should just contain only one entry for the most recently identified item.
- Do not include duplicate entries for the same nth_item in a single response.
- If you revise a previous selection, your "utterance" may briefly acknowledge the correction, but it must not disclose any item index or positional information.
\end{promptbox}

\subsection{\staticprotocol{} Protocol}


Since there is no interaction between the Director and Matcher under this protocol, we simulate the Director model by programmatically sending the Matcher a full item description at a time without actually prompting a Director LVLM. The Matcher system prompt is given below.

\begin{staticpromptbox}
### Task Overview

{Task Overview}

### Role Instructions

You are the Matcher. In this role, you must carefully analyze the Director's descriptions and choose the most likely match for each target item based on the visual items visible to you.

### Output Format Instruction

{Matcher Output Format Instruction}
\end{staticpromptbox}

\subsection{\fullprotocol{} Protocol}

\noindent \textbf{Director system prompt.}

\begin{fullpromptbox}
### Task Overview

{Task Overview}

### Role Instructions

You are the Director. In this role, you must describe each visual item visible to you from the lowest index to the highest index one at a time and answer any questions the Matcher may have.

Analyze the Matcher's question carefully based on your visual context and answer the question to the best of your ability.

### Communication Guidelines

- Describe exactly one target item at a time.
- Do not move on to the next item until the Matcher signals to proceed. Wait for them to signal when they are ready.
- Use natural, human-like, simple, and clear language that supports efficient communication.
- Try your best to make each description unambiguous and sufficiently informative for the Matcher to identify the related target item.
- Keep each description within 50 words, favoring shorter descriptions when clarity is maintained.
- Do not reference display position, display order, neighboring-item relations, or explicit item indices. You may describe within-item spatial structure when it is part of the item's appearance.
\end{fullpromptbox}

\noindent \textbf{Matcher system prompt.}

\begin{fullpromptbox}
### Task Overview

{Task Overview}

### Role Instructions

You are the Matcher. In this role, you must carefully analyze the Director's descriptions and ask any clarification questions if needed to identify each target item based on the visual items visible to you.

### Communication Guidelines

- In each turn, you are allowed to ask a question or signal to proceed to the next target item, but keep it no longer than 50 words. 
- If you ask a question or a confirmation, you must not internally select any target item until the Director provides an answer to your question. 
- If you signal to proceed, you must have internally selected a target item in your output. 
- Use concise, natural, human-like language and never mention item indices or positional cues in your natural-language utterance.
- You may ask as many questions as needed to clarify your understanding of the target items, but avoid asking redundant or irrelevant questions.
- Once you have identified a target item, you can move on to the next one until all the target items have been identified.
- You can ask questions about your previous selection(s), in case you want to double check or revise them.
- Do not reference display position, display order, neighboring-item relations, or explicit item indices. You may describe within-item spatial structure when it is part of the item's appearance.

### Output Format Instruction

{Matcher Output Format Instructions}
\end{fullpromptbox}

\subsection{\partialprotocol{} Protocol}

\noindent \textbf{Director system prompt.}

\begin{partialpromptbox}
### Task Overview

{Task Overview}

### Role Instructions

You are the Director. In this role, you must describe each visual item visible to you from the lowest index to the highest index one at a time and answer any questions the Matcher may have.

Analyze the Matcher's question carefully based on your visual context and answer the question to the best of your ability.

### Communication Guidelines

- Describe exactly one target item at a time.
- Do not move on to the next item until the Matcher signals to proceed. Wait for them to signal when they are ready.
- Use natural, human-like, simple, and clear language that supports efficient communication.
- For each new target, begin with an underspecified description to encourage the Matcher to ask clarifying questions.
- When responding to the Matcher's questions, be helpful and informative, but also be incremental and partial descriptions at a time, instead of over-sharing.
- Do not reference display position, display order, neighboring-item relations, or explicit item indices. You may describe within-item spatial structure when it is part of the item's appearance.
\end{partialpromptbox}

\noindent \textbf{Matcher system prompt.}

\begin{partialpromptbox}
### Task Overview

{Task Overview}

### Role Instructions

You are the Matcher. In this role, you must carefully analyze the Director's descriptions and ask any clarification questions if needed to identify each target item based on the visual items visible to you.

### Communication Guidelines

- In each turn, you are allowed to ask a question or signal to proceed to the next target item, but keep it no longer than 50 words. 
- If you ask a question or a confirmation, you must not internally select any target item until the Director provides an answer to your question. 
- If you signal to proceed, you must have internally selected a target item in your output. 
- Use concise, natural, human-like language and never mention item indices or positional cues in your natural-language utterance.
- You may ask as many questions as needed to clarify your understanding of the target items, but avoid asking redundant or irrelevant questions.
- Once you have identified a target item, you can move on to the next one until all the target items have been identified.
- You can ask questions about your previous selection(s), in case you want to double check or revise them.
- Do not reference display position, display order, neighboring-item relations, or explicit item indices. You may describe within-item spatial structure when it is part of the item's appearance.

### Output Format Instruction

{Matcher Output Format Instruction}
\end{partialpromptbox}

\subsection{\oracleprotocol{} Protocol}

\noindent \textbf{Director system prompt.}

\begin{oraclepromptbox}
### Task Overview

{Task Overview}

### Role Instructions

You are the Director. In this role, you must answer the Matcher's questions using only one of the following responses: "Yes," "No," or "Unclear."

Analyze the Matcher's question carefully based on your visual context and answer the question to the best of your ability.

### Communication Guidelines

- Just output "Yes," "No," or "Unclear" without any additional explanation or information in each turn.
- If the Matcher's response indicates the target item index they selected, just respond with "Unclear."
\end{oraclepromptbox}

\noindent \textbf{Matcher system prompt.}

\begin{oraclepromptbox}
### Task Overview

{Task Overview}

### Role Instructions

You are the Matcher. In this role, you must ask only yes or no questions to the Director to identify each target item based on the visual items visible to you. There are a total of \$num_targets target items for you to identify in each round.

### Communication Guidelines

- In each turn, you are only allowed to ask a clear and concise question at a time, no longer than 50 words.
- Avoid asking questions that require the Director to provide information beyond "Yes," "No," or "Unclear."
- Use concise, natural, human-like language and never mention item indices or positional cues in your natural-language utterance.
- You may ask up to \$max_questions questions as needed to clarify your understanding of the target items, and avoid asking redundant or irrelevant questions.
- Once you have identified a target item, you can move on to the next one until all the \$num_targets target items have been identified.
- You can ask questions about your previous selection(s), in case you want to double check or revise them.
- For each question, make clear which target item it refers to, such as "the first target item," "the fifth target item," and so on.

### Output Format Instruction

{Matcher Output Format Instruction}
\end{oraclepromptbox}

\section{Prompts for Referring Expression Processing}

\subsection{Referring Expression Extraction}

Prompt used for the automatic extraction of referring expressions for each target visual referent produced in a referential conversation. Words within ``<>" denote placeholders. 

This prompt is from \citet{zeng2026lvlmshumansgrounddifferently} and we only use it for the dog and basket datasets. The tangrams dataset already comes with manually curated object descriptions for each target object.

\begin{promptbox}
This is an extractive task.

You will be given a transcript of a conversation between two participants engaged in a collaborative object-matching task. There are exactly <num_objects> target objects. One participant (the describer) describes each target object, and the other participant (the matcher) attempts to identify them.

Your task is to extract the descriptive phrases used by the describer for each target object. 
- Extract phrases verbatim from the transcript. 
- Do not extract the whole utterance, only the descriptive phrases.
- Exclude disfluencies, fillers, and false starts (e.g., "um", "uh", "like").
- Do not paraphrase or infer missing information.
- Each object may have one or multiple descriptive phrases.

Return the results in the following JSON format:

{
    "object_#1": "descriptive phrases for object 1",
    "object_#2": "descriptive phrases for object 2",
    ...
    "object_#<num_objects>": "descriptive phrases for object <num_objects>"
}

Example description phrases:

- doesn't have handle, tip of it is thicker than rest of body, brownish color, weaves are in squares if you look at it directly
- half circle, no handles, top tip of it is a little bit thicker than rest of body
- tip which is a little bit thicker than rest of body
- tip that is a little bit larger than body, looks a little bit thicker

Transcript:

<transcript>

Output only the JSON object. Do not include any additional text or explanations.
\end{promptbox}

\subsection{Partial Description Extraction}

\begin{promptbox}
Given a description map in JSON format, convert it into a new JSON object with the same top-level keys.

1. "full": the original full description as a string, unchanged.
2. "attributes": a list of atomic attributes extracted from the full description.
3. "least": the least informative attribute for this visual item, selected from its "attributes" list.

Attribute extraction rules:
- Each attribute should be expressed in natural language.
- Each attribute must describe only one visual feature.
- Keep attributes as close as possible to the original wording; verbatim if possible.
- Do not infer attributes that are not explicitly stated in the description.
- Do not treat missing attributes as negative evidence.
- Remove duplicate attributes within the same item.
- Split compound descriptions into separate atomic attributes.
- Preserve relative/comparative attributes when explicitly stated, such as "tallest", "shortest", "second shortest", or "shorter than 9".
- Resolve coreferences when extracting attributes, so that each attribute is self-contained and does not rely on pronouns or other references to other attributes.
- If there is only one attribute in the description, simply reuse the full description as the single attribute.

Least attribute selection rules:
- For each item, choose exactly one attribute from that item's "attributes" list.
- Choose the attribute that is least informative for identifying the item, taking into account all attributes across all items in the input map.
- Do not choose an attribute that gives the gestalt impression of the whole item; instead, choose an attribute that describes a specific visual feature.
- Prefer a generic, common, or weakly distinguishing attribute over an attribute that is distinctive or rare across items.
- If several attributes are similarly generic, choose the one that would be least helpful on its own for distinguishing the item from the other items.
- Do not invent, rewrite, or combine attributes for "least"; it must exactly match one string in that item's "attributes" list.

Return only valid JSON. Do not include explanations, markdown, or comments.

Input format:

{
  "image_id": "full natural-language description"
}

Output format:
{
  "image_id": {
    "full": "original full natural-language description",
    "attributes": [
      "attribute 1",
      "attribute 2"
    ],
    "least": "attribute 2"
  }
}
\end{promptbox}

\subsection{Object Description Generation}

We use the following prompt to generate full descriptions for each object across the four image sets with GPT-5.4. To mirror the human Director setting, the model is shown only the target images. We prompt it to produce a description map that can be used directly in our experimental pipeline.  

\begin{promptbox}
You are creating a full description map for a visual reference task.

The task involves two roles: a Director and a Matcher. The Director sees a set of target visual items and must describe them so the Matcher can identify the same items among possible distractors. Your job is to write the Director-style initial descriptions for one round.

You will receive all director images for the round in a single message. Each image is labeled with its filename only so the output can be keyed correctly.

Return only valid JSON: an object mapping each exact filename to one short natural-language description.

Description rules:
- Describe every provided image exactly once.
- Use the exact filenames as JSON keys.
- Each description must be fewer than 50 tokens.
- Use natural, human-like, simple, and clear language.
- Make each description unambiguous and sufficiently informative for a Matcher to identify the item among similar items.
- Describe the item itself: shape, pose, orientation, silhouette, distinctive parts, and within-item spatial structure when useful.
- Do not reference display position, display order, neighboring-item relations, explicit item indices, file paths, or filenames in the descriptions.
- Do not say "image", "picture", "item number", or similar display-oriented labels in the descriptions.
- Do not include markdown, comments, explanations, or extra keys.
\end{promptbox}

\section{Dataset Details\label{app:datasets}}

\begin{figure*}[!htb]
    \centering    \includegraphics[width=0.95\linewidth]{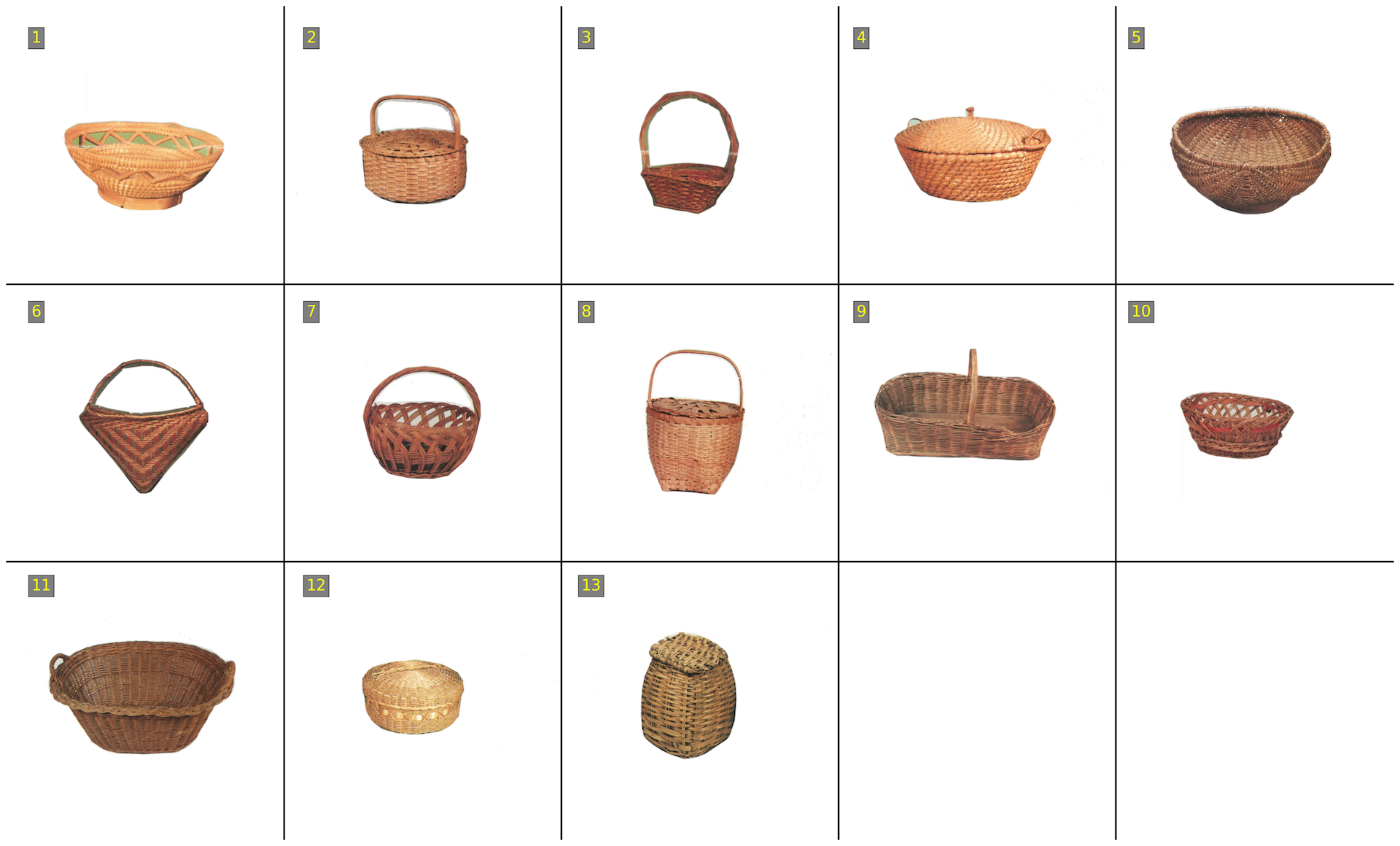}
    \caption{Basket image set from \citet{wang-etal-2025-lvlms}. The first 10 baskets are target items visible to the Director; the Matcher sees the full set.}
    \label{fig:baskets_emnlp2025}
\end{figure*}

\begin{figure*}[]
    \centering    \includegraphics[width=0.95\linewidth]{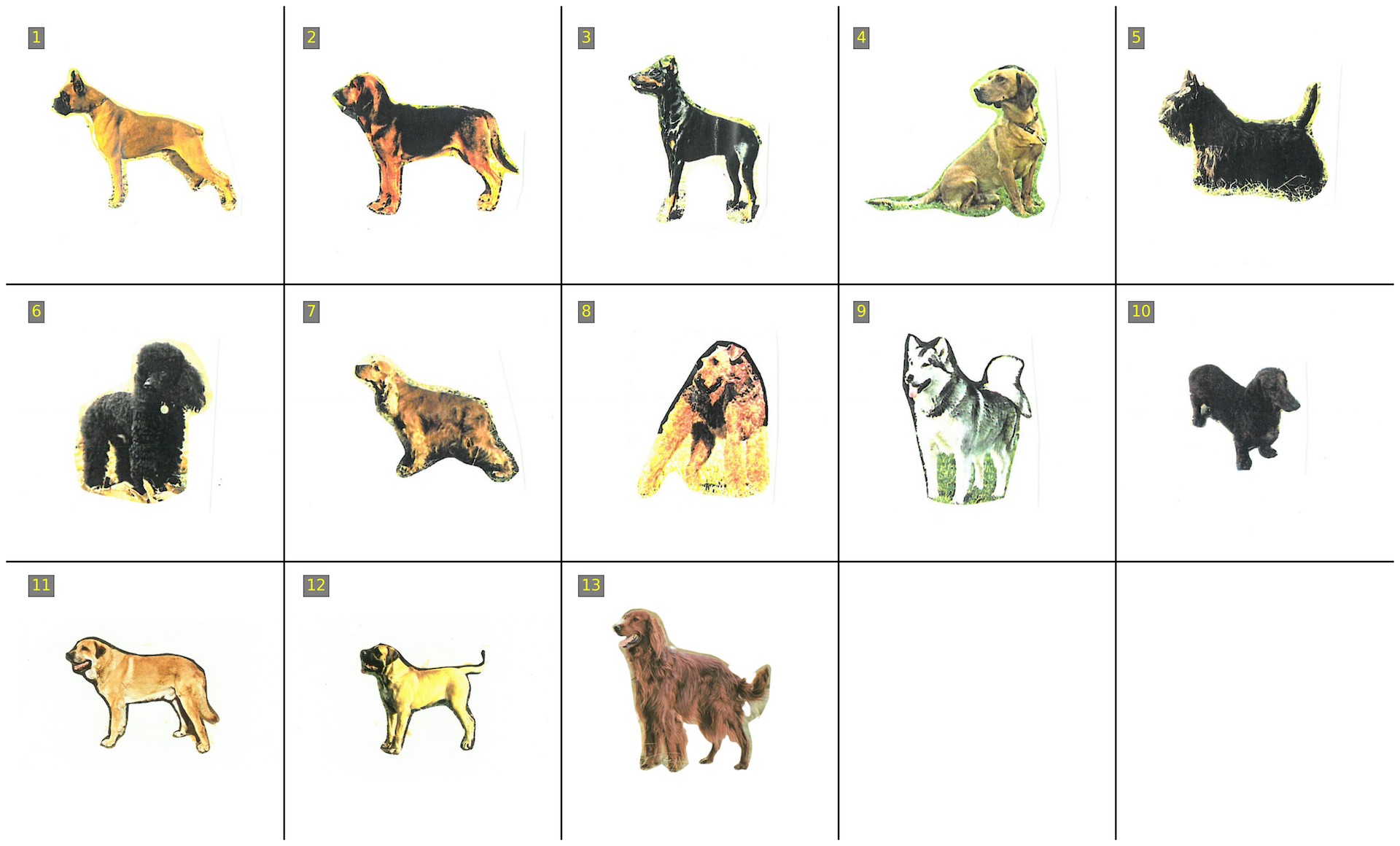}
    \caption{Dog image set from \citet{wang-etal-2025-lvlms}. The first 10 dogs are target items visible to the Director; the Matcher sees the full set.}
    \label{fig:dogs_emnlp2025}
\end{figure*}

\begin{figure*}[]
    \centering    \includegraphics[width=0.95\linewidth]{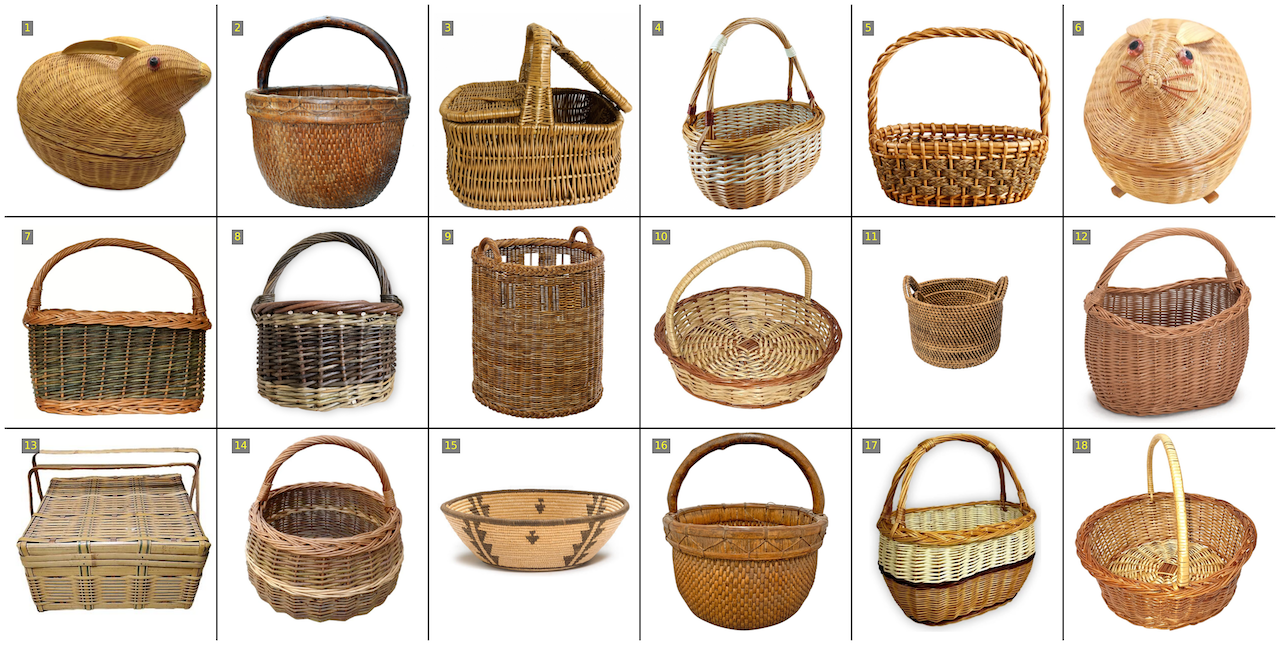}
    \caption{Basket image set from \citet{zeng2026lvlmshumansgrounddifferently}. The first 12 baskets are target items visible to the Director; the Matcher sees the full set.}
    \label{fig:baskets_acl2026}
\end{figure*}

\begin{figure*}[]
    \centering    \includegraphics[width=0.95\linewidth]{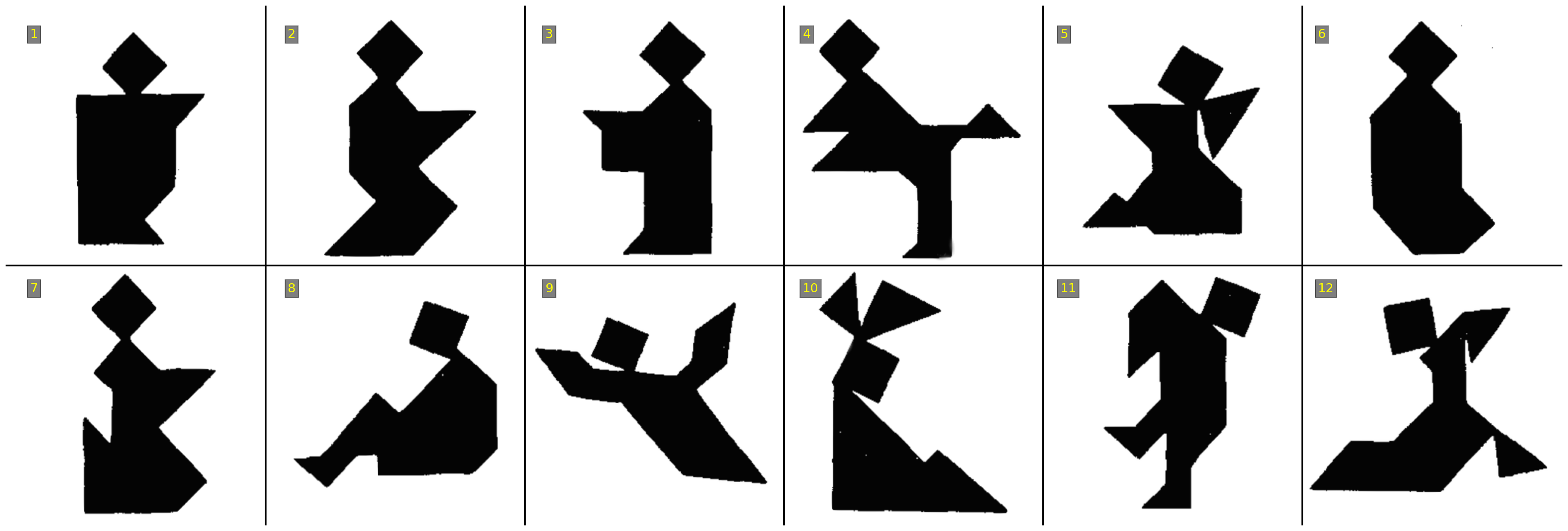}
    \caption{Tangram image set from \citet{Hawkins2020}. Both the Director and Matcher see the full set; item orders are randomized independently for the two views in our experiments.}
    \label{fig:tangrams_hawkins2020}
\end{figure*}

Figures~\ref{fig:baskets_emnlp2025},~\ref{fig:dogs_emnlp2025},~\ref{fig:baskets_acl2026}, and~\ref{fig:tangrams_hawkins2020} show the four object image sets used in our experiments. These sets span real-world objects (dogs and baskets) and abstract tangrams, and differ in the number of target and distractor items. For the basket and dog sets, the Director sees only the target items, while the Matcher sees the full candidate set. For the tangram set, both players see the full set. In all experiments, item orders for Director follow the item-description order in the original dataset, while Matcher's item orders are randomized to prevent trivial index matching or data contamination.

We obtain the image sets and corresponding human--human interaction data from the GitHub repositories associated with the cited papers.

\end{document}